\documentclass[conference]{IEEEtran}
\usepackage{times}
\usepackage[numbers]{natbib}
\usepackage{multicol}
\usepackage[bookmarks=true]{hyperref}
\usepackage{xspace}
\usepackage{booktabs}
\usepackage{amsmath}
\usepackage{graphicx}
\usepackage{wrapfig}
\usepackage{amssymb}
\usepackage{listings}
\usepackage[dvipsnames]{xcolor}
\usepackage{utfsym}
\usepackage{colortbl}
\usepackage{caption}
\usepackage{arydshln}

\usepackage[capitalize]{cleveref}
\crefname{section}{Sec.}{Secs.}
\Crefname{section}{Section}{Sections}
\Crefname{table}{Table}{Tables}
\crefname{table}{Tab.}{Tabs.}
\crefname{appendix}{App.}{Apps.}
\Crefname{appendix}{Appendix}{Appendices}

\renewcommand{\authorrefmark}[1]{\textsuperscript{#1}}
\definecolor{mycolor}{rgb}{0.07,0.419,0.682}
\definecolor{txcolor}{rgb}{0.07,0.419,0.682}
\newcommand{\red}[1]{\textcolor{black}{#1}}
\hypersetup{
    colorlinks=true,
    linkcolor=mycolor, 
    urlcolor=mycolor,  
    citecolor=mycolor, 
}

\newcommand{\ci}[1]{\tiny{\textcolor{gray}{~($\pm #1$)}}}

\definecolor{lightgray}{RGB}{250,250,250} 
\newcommand{\ourrow}{\rowcolor{lightgray}\bfseries}

\newcommand{\ourshort}{HOMIE\xspace}
\newcommand{\yes}{\large \color{OliveGreen}\checkmark}
\newcommand{\no}{\color{BrickRed} \scalebox{1}{\usym{2613}}}
\begin{document}

\title{HOMIE: Humanoid Loco-Manipulation \\ with Isomorphic Exoskeleton Cockpit}


\author{\authorblockN{Qingwei Ben\authorrefmark{1,2,*},
Feiyu Jia\authorrefmark{1,*},
Jia Zeng\authorrefmark{1}, 
Junting Dong\authorrefmark{1},
Dahua Lin\authorrefmark{1,2},
Jiangmiao Pang\authorrefmark{1}\\
\authorblockA{\authorrefmark{1} Shanghai AI Laboratory}
\authorblockA{\authorrefmark{2} Multimedia Laboratory, The Chinese University of Hong Kong}}
\authorblockA{\authorrefmark{*} Authors with equal contribution}}
\setlength{\parskip}{0pt} 
\let\oldtwocolumn\twocolumn
\renewcommand\twocolumn[1][]{%
    \oldtwocolumn[{#1}{
    \begin{flushleft}
           \centering
    \vspace{-9pt}
           
    \includegraphics[clip,trim=0cm 0cm 0cm 0cm,width=1.0\textwidth]{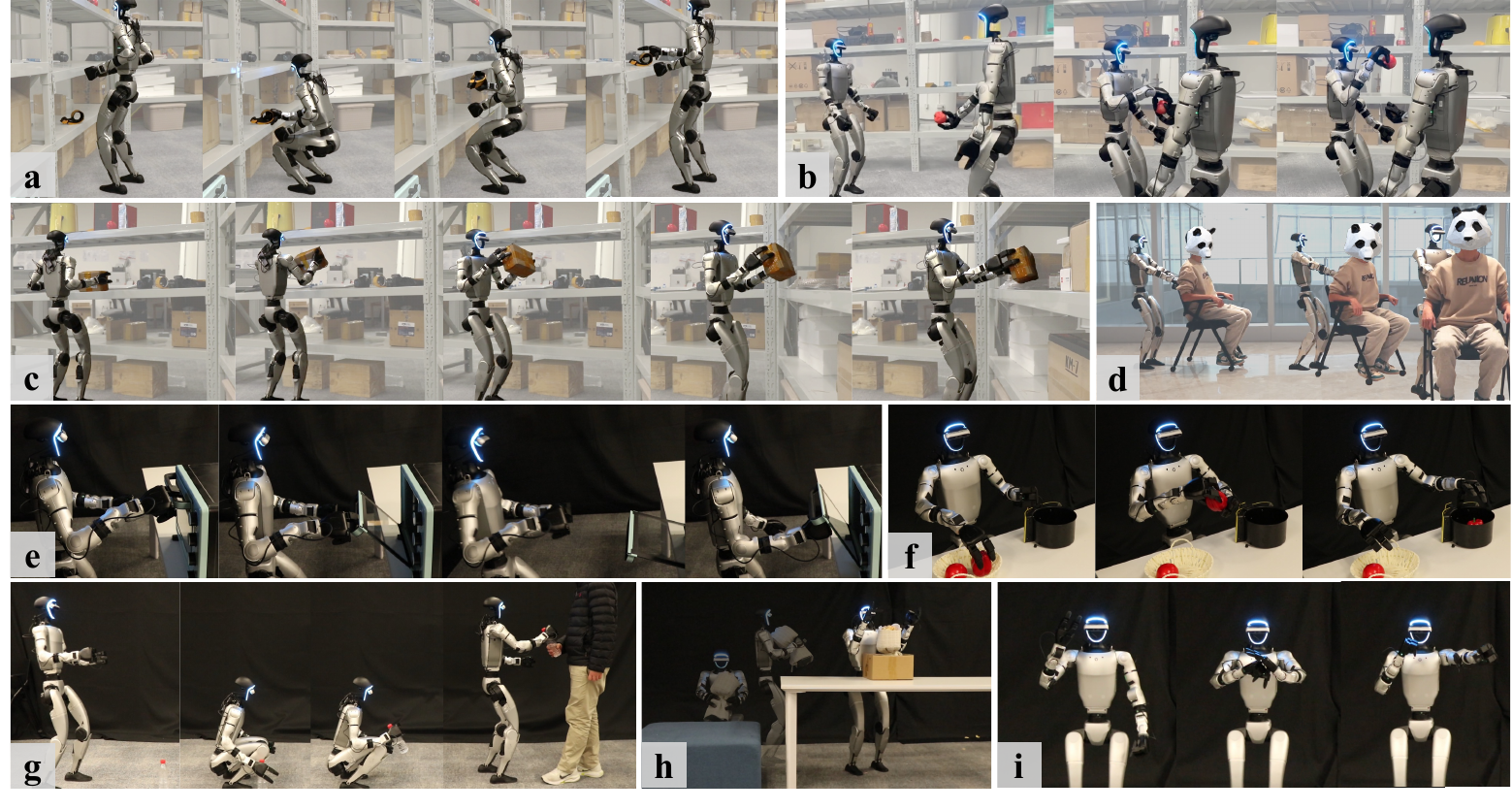}\captionsetup{justification=raggedright,singlelinecheck=false}
    \captionof{figure}{
        \ourshort empowers the humanoid robot to execute various loco-manipulation tasks in the real world. \textcolor{mycolor}{(a):} Squatting to grasp a tape and placing it on a higher shelf; \textcolor{mycolor}{(b):} Facilitating an apple handover between two robots; \textcolor{mycolor}{(c):} Holding a box and transferring it to another shelf; \textcolor{mycolor}{(d):} Pushing a person seated on a chair; \textcolor{mycolor}{(e):} Opening an oven door; \textcolor{mycolor}{(f):} Picking up a tomato, handing it over, and placing it in a fruit basket; \textcolor{mycolor}{(g):} Retrieving a bottle from the ground; \textcolor{mycolor}{(h):} Holding a flower and placing it into a box on a table. \textcolor{mycolor}{(i)}: Balancing under vigorous motion. These tasks show robustness and generality of~\ourshort.
    }
    \label{fig:teaser}
    \end{flushleft}
    }]
    
}

\maketitle

\begin{abstract}

    \red{Generalizable humanoid loco-manipulation poses significant challenges, requiring coordinated whole-body control and precise, contact-rich object manipulation. To address this, this paper introduces HOMIE, a semi-autonomous teleoperation system that combines a reinforcement learning policy for body control mapped to a pedal, an isomorphic exoskeleton arm for arm control, and motion-sensing gloves for hand control, forming a unified cockpit to freely operate humanoids and establish a data flywheel.
    The policy incorporates novel designs, including an upper-body pose curriculum, a height-tracking reward, and symmetry utilization. These features enable the system to perform walking and squatting to specific heights while seamlessly adapting to arbitrary upper-body poses.
    The exoskeleton, by eliminating the reliance on inverse dynamics, delivers faster and more precise arm control. 
    The gloves utilize Hall sensors instead of servos, allowing even compact devices to achieve 15 or more degrees of freedom and freely adapt to any model of dexterous hands.
    Compared to previous teleoperation systems, HOMIE stands out for its exceptional efficiency, completing tasks in half the time; its expanded working range, allowing users to freely reach high and low areas as well as interact with any objects; and its affordability, with a price of just \$500.}
    The system is fully open-source, demos and code can be found in our \href{https://homietele.github.io/}{website}.
\end{abstract}

\IEEEpeerreviewmaketitle

\section{Introduction}
\red{Generalizable humanoid loco-manipulation is crucial for integrating humanoid robots into daily life and enabling them to handle labor-intensive tasks. Achieving this requires coordinated whole-body control (WBC) policies that endow robots with both strong athletic capabilities and precise, contact-rich object manipulation skills for interacting with a variety of objects. Teleoperation is a promising technique to realize this vision, leveraging a data-driven approach to create a flywheel effect. However, the field currently faces a significant dichotomy: reinforcement learning (RL)-trained locomotion policies excel at environmental adaptation but lack the interfaces needed for real-time, precise teleoperation \cite{fu2024humanplus,he2024learning,he2024omnih2o,cheng2024express,ji2024exbody2,lu2024pmp}. On the other hand, most existing teleoperation systems focus solely on upper-body control without considering the impact of locomotion on the robot's operational workspace, thereby severely limiting its functionality \cite{cheng2024open,yang2024ace,ze2024humanoid_manipulation,qin2023anyteleop,iyer2024open}. This fragmentation creates a lose-lose situation, where robots either sacrifice dexterous manipulation during movement or compromise their workspaces when performing manipulation.}

\red{The path forward demands mutual perspective iteration: RL-based training incorporates upper-body teleoperation interfaces without compromising the robot's athletic ability, and teleoperation system seamlessly integrate locomotion control modules while affording accurate and smooth pose acquisition. In responce, we introduce \ourshort, a semi-autonomous humanoid teleoperation system that integrates a RL policy for body control mapped to a pedal, an isomorphic exoskeleton arm for arm control, and motion-sensing gloves for hand control. This unified cockpit enables a single operator to precisely and efficiently control a humanoid robot's full-body movements, addressing both humanoid whole-body control and real-time precise teleoperation.}

\red{Our RL-based training framework features three core techniques: upper-body pose curriculum for dynamic balance adaptation, height-tracking reward for precise squatting, and symmetry utilization for action regularization and data augmentation. These components collectively enhance the robot's physical agility, enabling robust walking, rapid squatting to any required heights, and stable balance maintenance during dynamic upper-body movements, thereby significantly expanding the robot's operational workspace beyond existing solutions and allowing any teleoperation commands to take effect. Unlike previous whole-body control methods that depend on motion priors derived from MoCap data~\cite{AMASS}, our framework eliminates this dependency, resulting in a more efficient pipeline. }

\red{Complementing the training framework, our hardware system features isomorphic exoskeleton arms, a pair of motion-sensing gloves, and a pedal. The pedal design serves as an effective interface for locomotion command acquisition, guiding the robot's movement while freeing the operator's upper body. This setup enables simultaneous acquisition of upper-body poses and removes the need for continuous synchronized walking between the operator and the robot. To eliminate the inaccuracies introduced by inverse kinematics (IK) and pose estimation, which are commonly used in mainstream teleoperation systems, we design the exoskeleton arms to be isomorphic to the controlled robot. This allows us to directly set the upper-body joint positions based on the exoskeleton readings, bypassing the need for IK and resulting in faster and more accurate teleoperation. Each of our gloves offers 15 degrees of freedom (DoF), surpassing most existing dexterous hands, enabling them to control a variety of hand types using the same gloves. Additionally, the gloves can be detached from the arms, making them reusable in systems isomorphic to different robots. The total cost of the hardware system is just \$0.5k, significantly more affordable than motion capture (MoCap) devices~\cite{wang2024dexcap}.}

\red{Through ablation experiments, we validate the effectiveness of each technique in our training framework and demonstrate the robustness of the resulting policies across different robots. Our evaluation shows that the hardware system supports 200\% faster and more accurate pose acquisition than previous methods, enabling operators to complete tasks more efficiently than virtual reality (VR)-based approaches. Real-world studies confirm that the trained policies can be deployed directly in the real world, allowing robots to perform diverse loco-manipulation tasks stably in complex environments. We further show that real-world data collected via~\ourshort can be effectively used by imitation learning (IL) algorithms, allowing humanoid robots to autonomously execute tasks. Integrated into simulation environments, our cockpit also enables seamless teleoperation in virtual settings. }

\red{In summary, the key contributions of~\ourshort are:}
\begin{enumerate}
    \item \red{A novel humanoid teleoperation cockpit that combines RL-based loco-manipulation control with an isomorphic exoskeleton and motion-sensing gloves, enabling full-body control by a single operator.}
    \item \red{The first successful implementation of teleoperation-compatible humanoid loco-manipulation, including dynamic squatting, without relying on motion prior data.}
    \item \red{A cost-effective hardware system that supports more precise and faster whole-body control than existing systems, significantly reducing task completion times.}
\end{enumerate}

\section{Related Works}
\begin{table*}[!ht]
\renewcommand{\arraystretch}{1.2}
\vspace{-0.15in}
    \centering
    \caption{\textbf{Comparison between representative teleoperation systems and~\ourshort.} \textcolor{txcolor}{Cost:} total cost of each system. \textcolor{txcolor}{Arm and Dex-Hand Tracking:} method of tracking arm and hand poses. \textcolor{txcolor}{Loco-Manip.:} whether or not have loco-manipulation capability. \textcolor{txcolor}{Whole-body:} whether or not teleoperate the whole body of humanoid robots. \textcolor{txcolor}{No Mocap:} whether or not exclude MoCap data.}
    \label{tab:teleop_comp}
    \resizebox{.96\textwidth}{!}{
    \begin{tabular}{l|cccccc}
        \toprule
        \textbf{Teleop System} & Cost (\$) & Arm Tracking  & Dex-Hand Tracking & Loco-Manip. & Whole-body & No MoCap  \\
        \midrule
        Mobile-ALOHA~\cite{fu2024mobile} & 32k & Joint-matching & \no & \yes & \no & \yes \\
        GELLO~\cite{wu2023gello} & 0.6k & Joint-matching & \no & \no & \no & \yes \\
        AirExo~\cite{fang2024airexo} & 0.6k & Joint-matching & \no & \no & \no & \yes \\
        ACE~\cite{yang2024ace} & 0.6k & Joint-matching &  Vision Retarget & \no & \no & \yes \\
        DexCap~\cite{wang2024dexcap} & 4k & Vision Retarget & Mocap + SLAM & \no & \no & \yes \\ 
        AnyTeleop~\cite{qin2023anyteleop}  & $\sim$ 0.3k & Vision Retarget & Vision Retarget & \no & \no & \yes\\ 
        OpenTelevision~\cite{cheng2024open} & 4k & VR devices & VR devices & \no & \no & \yes \\
        HumanPlus~\cite{fu2024humanplus} & 0.05k & Vision Retarget & Vision Retarget & \yes & \yes & \no \\
        OmniH2O~\cite{he2024omnih2o} & 0--3.5k & Vision / VR & Vision / VR & \yes & \yes & \no \\
        Mobile-TeleVision~\cite{lu2024pmp} & 3.5k & VR devices & VR devices & \yes & \yes & \no \\
        \midrule
        \ourrow \ourshort (Ours) & 0.5k & Joint-matching & Joint-matching & \yes & \yes & \yes \\
        \bottomrule
    \end{tabular}}
    \vspace{-1.4pt}
\end{table*}
\subsection{Teleoperation Systems}
Teleoperating dual-arm robots to perform complex manipulation tasks is an efficient way to collect real-world expert demonstration, which can then be used by IL to learn autonomous skills~\cite{fu2024mobile,cheng2024open,ze2024humanoid_manipulation,black2024pi_0,lin2024learning}. Some researchers utilize robotic arms identical to the teleoperated ones~\cite{zhao2023learning,fu2024mobile,wu2023gello,fang2024airexo}, making joint-matching possible, thus ensuring high accuracy and fast response speed. However, due to the high cost of robotic arms, the establishment of such a system incurs significant expenses. Additionally, teleoperating dexterous hands with these systems is not feasible. 

An alternative approach is to use VR devices~\cite{cheng2024open,iyer2024open} or just a camera~\cite{Sivakumar-RSS-22,qin2023anyteleop,li2024okami}. These works use vision-based techniques to capture the operator's wrist postures and key points of the hands, which are used by IK to calculate the joint positions of the arms and hands. However, due to limitations in the accuracy, inference speed, and difficulty in handling occlusions of pose estimation, such approaches cannot guarantee rapid and accurate pose acquisition. Some researchers try to use MoCap methods~\cite{wang2024dexcap,caeiro2021systematic,liu2019high,liu2017glove} to acquire more accurate poses at higher frequencies, but MoCap equipment is very expensive. Moreover, since IK is an iterative method that approximates solutions, even when wrist and hand poses are captured accurately, the limitations of IK may prevent the robot from achieving the desired posture.  Another possible solution is an exoskeleton-based teleoperation system, which does not require an additional identical robot, thus the overall cost is relatively low. Some research calculates the end-effector pose of the exoskeleton using Forward Kinematics (FK) and then apply IK to determine the robot's joint positions, while using computer vision techniques to capture the hand poses~\cite{yang2024ace}. However, these systems are also limited by the inaccuracies of IK and pose estimation. 

Some studies utilize isomorphic exoskeletons~\cite{wu2023gello,fang2024airexo}, which can also employ joint-matching to teleoperate the robots, ensuring both low cost and high accuracy and control frequency. Neverthless, these systems typically handle robotic arms equipped with grippers, limiting their application to basic manipulation tasks rather than dexterous ones. Since some projects have introduced cheap and reliable motion-sensing gloves~\cite{Nepyope2023Project-Homunculus,dafarra2024icub3}, redesigning and combining them with an exoskeleton could potentially overcome this limitation, a solution that has not yet been realized in this field. \ourshort is designed to combine all the advantages mentioned above, integrating isomorphic exoskeleton arms with a pair of novel motion-sensing gloves. We will introduce this system in \cref{sec:hard_design}. A comparison between \ourshort and previous representative teleoperation systems can be found in \cref{tab:teleop_comp}.

\subsection{Whole-body Loco-Manipulation}

To enable robots to perform whole-body loco-manipulation tasks, some researchers focus on model-based optimization algorithms~\cite{miura1984dynamic,wensing2023optimization,moro2019whole,zhang2024whole,chignoli2021humanoid}, particularly generating locomotion control laws by solving optimal control problems (OCPs). Despite significant efforts to make OCPs computationally tractable, these algorithms still struggle with complex scenarios due to their high computational demands during online processing. Reinforcement Learning (RL)-based algorithms, especially those based on Proximal Policy Optimization (PPO)~\cite{schulman2017proximal}, offer a more powerful alternative. Using these methods, several studies successfully achieve whole-body loco-manipulation in quadruped robots~\cite{liu2024visual,pan2024roboduet,portela2024whole,ha2024umi}, and some teach humanoid robots to traverse various terrains~\cite{gu2024advancing,long2024learninghumanoidlocomotionperceptive,chen2024learning,dugar2024learning,li2024reinforcement,li2021reinforcement,liao2024berkeley,zhang2024wococo} or perform parkour~\cite{zhuang2024humanoid}. 

Achievements in quadrupeds motivate researchers to apply the same techniques to humanoid whole-body loco-manipulation~\cite{gu2025humanoid}. Some studies train whole-body policies for humanoid robots~\cite{ji2024exbody2}, enabling them to act in a manner similar to human operators or even dance with people. Some other research separates the upper and lower body~\cite{cheng2024express,fu2024humanplus,he2024learning,he2024omnih2o,lu2024pmp}, using policies trained by RL to control the lower body while directly setting the joint positions of the upper body, thus helping robots achieve better balance. 
Despite achieving impressive results, these methods still face several common limitations. First, they often rely on retargeted MoCap data~\cite{AMASS} to get motion prior~\cite{luo2023universal} for training robots. However, obtaining MoCap data is costly, and adapting robots to new poses necessitates additional data collection, which significantly hinders the scalability of these approaches. Second, many of these methods employ vision-based algorithms to estimate the operator's poses, which lack the precision of exoskeleton-based devices. This limitation reduces the accuracy required for humanoid robots to perform loco-manipulation tasks effectively. Third, these methods generally fail to incorporate the ability to control a robot's body height. Height control is crucial for handling objects at varying elevations, and its absence severely restricts the robot's operational workspace. Finally, when issuing locomotion commands, some studies rely on body movement data directly~\cite{cheng2024express,fu2024humanplus,he2024learning,he2024omnih2o}, while others use joysticks or pedals~\cite{lu2024pmp}. The former approach becomes impractical when operators need to control robots in large-scale environments, whereas the latter offers a more effective solution. However, controlling with joysticks necessitates the use of hands, which may already be occupied by other manual tasks, thereby highlighting the advantage of using pedals for locomotion commands.

\begin{figure*}[!ht]
  \centering
  \includegraphics[width=1.00\textwidth]{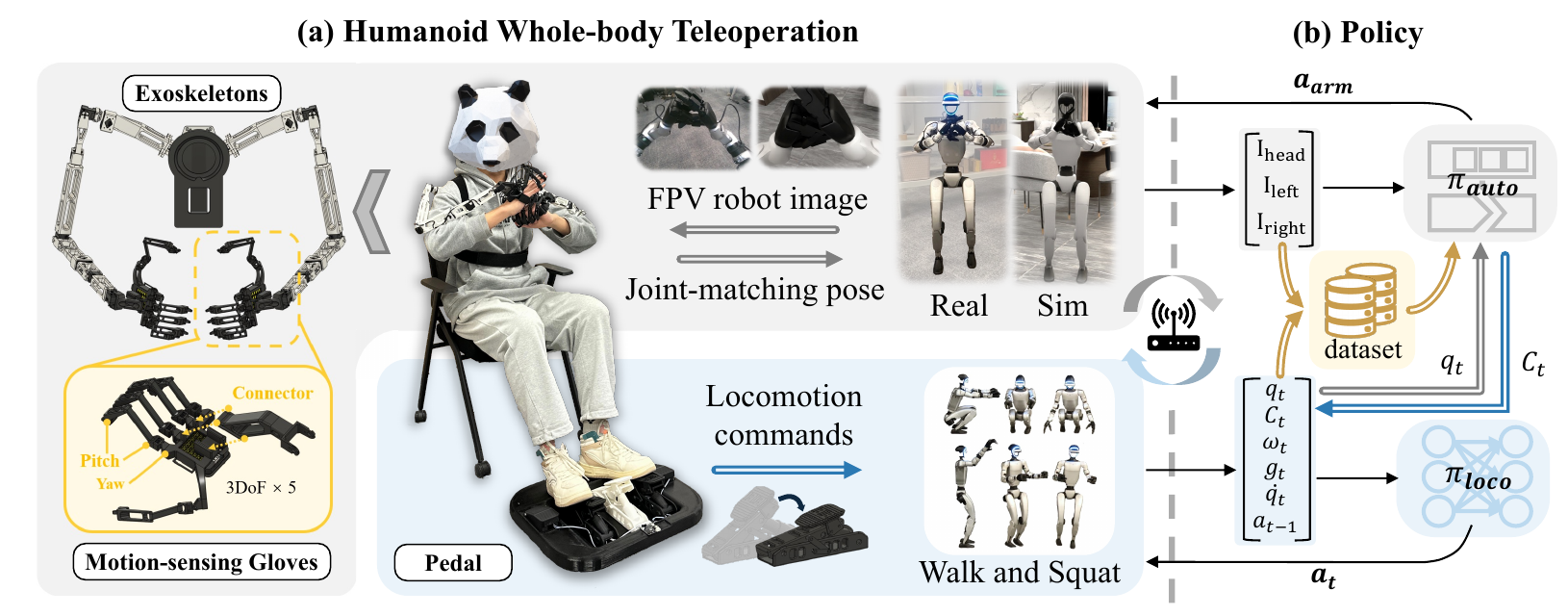}
  \caption{\textbf{System Overview.} \textcolor{mycolor}{(a):} how an operator uses the exoskeleton-based hardware system to control humanoid robots in the real world and simulation. \textcolor{mycolor}{(b):} how $\pi_{loco}$ controls the robots, the data collection process for training $\pi_{auto}$, and how $\pi_{auto}$ takes over the operator to control the robots. Communication between the cockpit and the robot is achieved via Wi-Fi.}
  \label{fig:overview}
\end{figure*}
\section{Method}
\label{sec:method}
\subsection{System Overview}
As shown in \cref{fig:overview}, \ourshort consists of a low-level policy $\pi_{loco}$ and an exoskeleton-based hardware system. At any given time $t$, \red{the first point of view (FPV) of the robot will be transferred by Wi-Fi to the display inside the cockpit, so the operator can teleoperate the robot with FPV}. By stepping on the pedal, the operator provides the required locomotion commands \textbf{$C_t=[v_{x,t}, \omega_{yaw,t}, h_t]$} where $v_{x,t}$ is the desired forward or backward speed, $\omega_{yaw,t}$ is the turning speed, and $h_t$ is the target height of the robot's torso. The policy $\pi_{loco}$ controls the robot's lower-body based on $C_t$. Meanwhile, the operator controls the exoskeleton to provide the required joint angles $q_{upper}$ for the robot's upper-body, which are directly set to the robot. The upper and lower bodies work in coordination, continuously cycling through the process, ultimately enabling teleoperating robots to complete loco-manipulation tasks either in the real world or in simulation. Communications between the cockpit and the robot are achieved via Wi-Fi, allowing operation even when the robot is far from the hardware system. We can collect demonstrations while teleoperating the robot and use them to train an autonomous policy $\pi_{auto}$. Once trained successfully, $\pi_{auto}$ can take over the operator to give $C_t$ and $q_{upper}$, thus driving the robot to perform tasks autonomously.

\subsection{\red{Humanoid Whole-body Control}}
\label{sec:method_rl}
\begin{figure}[!ht]
  \centering
  \includegraphics[width=0.49\textwidth]{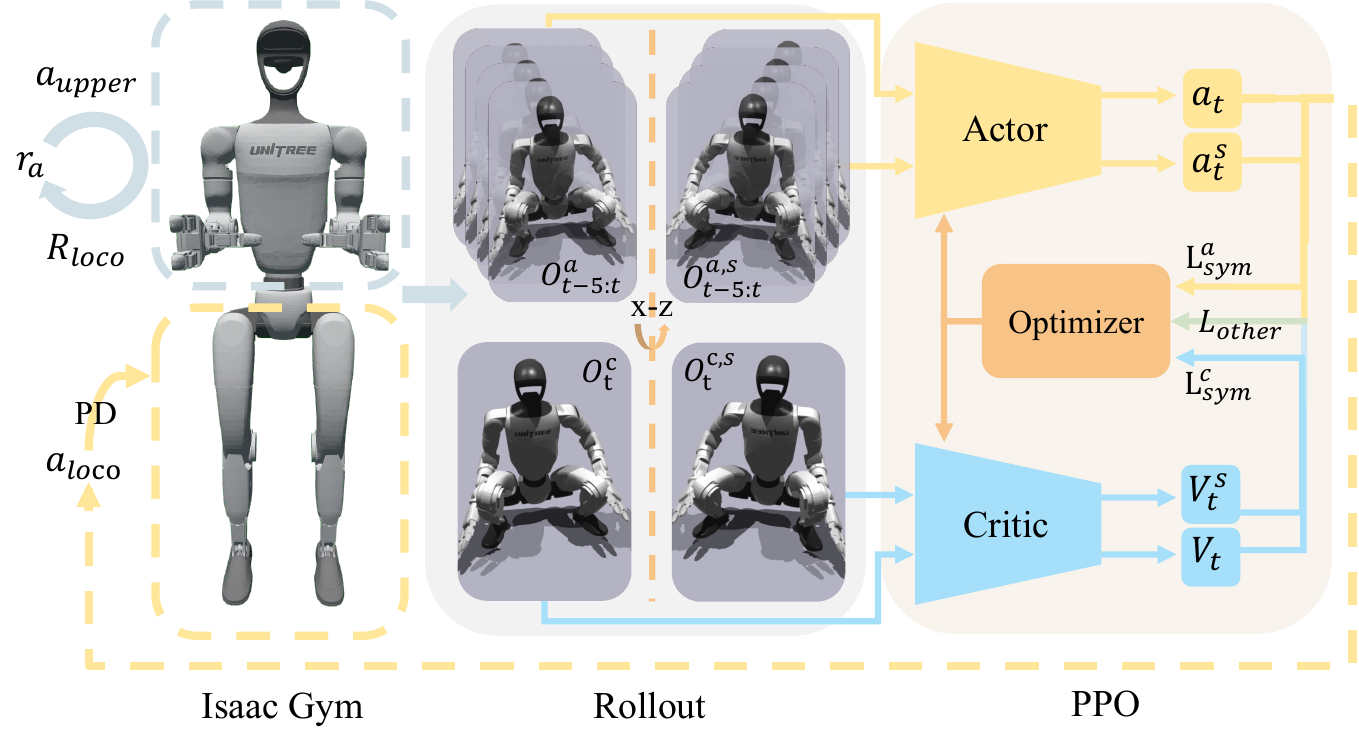}
  \caption{RL training framework of~\ourshort.}
  \label{fig:rl}
\end{figure}
To enable humanoid robots to perform loco-manipulation tasks, we design an RL-based training framework, which trains different robots to accomplish squatting and walking under continuously changing upper-body poses. We take Unitree G1 as an example and show the process of the framework in \cref{fig:rl}. The policy $\pi_{loco}$ trained by this process is capable of zero-shot sim-to-real transfer. We introduce the training settings and three key techniques of our framework in this section.

\subsubsection{Training Settings}
The observations of one step are defined as \textbf{$O_t=[C_t,\,\omega_t,\,g_t,\,q_t,\,\dot{q}_t,\,a_{t-1}]$}, where $C_t$ is the command, $\omega_t$ is the body's angular velocity, $g_t$ is the projection of $\vec{g}=[0,0,-1]$ in the robot's torso coordinate frame, $q_t$ is the joint angles of all joints of the robot, $\dot{q}_t$ is the joint velocities of all joints of robot, $a_{t-1}$ is the last time action. Then we can get the whole observations of $\pi_{loco}$ by concatenating $O_{t-5:t}$. The actions $a_t$ of the policy correspond one-to-one with the joints of the robot's lower body. After the neural network computes $a_t$ based on $O_{t-5:t}$, we use 
\begin{equation}
\tau_{t,i} = Kp_i \times (a_{t,i}-q_{0,t,i}) - Kd_i \times \dot{q_{t,i}} 
\end{equation}
to calculate the torques for joint motors, thereby driving the motors to work and enabling the robot's movement. In the equation, $i$ denotes the index of joints, $\{Kp_i\}$ and $\{Kd_i\}$ are stiffness and damping of each joint, $\{q_{0,t,i}\}$ are default joint positions of each joint. Our framework is implemented based on the code of ~\cite{long2024hybrid,rudin2022learning}, and more training details can be found in Appendix \ref{appendix:RL}.
\subsubsection{Upper-body Pose Curriculum}
We use a curriculum learning technique to ensure that $\pi_{loco}$ can still complete locomotion tasks under any continuously varying poses of the robot's upper-body. We adjust the sampling range of the upper body joint angles using the upper action ratio $\rho_a$. At the start of training, $r_a$ is set to 0. Each time the policy drives the robot to track the linear velocity with a reward function that reaches the threshold, \red{$\rho_a$} increases by 0.05, eventually reaching 1. We first sample \red{$\rho_a'$} from the probability distribution
\begin{equation}
    \red{p(\rho_a'|\rho_a) = \frac{20(1 - \rho_a) \, e^{-20(1 - rho_a)\rho_a'}}{1 - e^{-20(1 - \rho_a)}},\,\,\,\rho_a\in [0, 1)}
\end{equation}
and then resample $a_i$ by \red{$\mathcal{U}(0, \rho_a')$}. We actually sample $a_i$ by
\begin{equation}
\label{equa:sample}
    \red{a_i = \mathcal{U}(0, -\frac{1}{20 (1 - \rho_a)} \ln\left(1 - \mathcal{U}(0, 1)\left(1 - e^{-20 (1 - \rho_a)}\right)\right))}
\end{equation}
As $r_a$ increases, the probability distribution gradually transitions from being close to 0 to $\mathcal{U}(0,1)$. This ensures that during the curriculum process, the probability distribution consistently satisfies \red{$p(\rho_a'|\rho_a)>0,\forall \rho_a'\in[0,1],\rho_a\in[0,1)$}. Compared to directly using \red{$\mathcal{U}(0,\rho_a)$}, this method approaches the final target in a more gradual and smoother manner. For better understanding of \cref{equa:sample}, we visualize it in Appendix \ref{appendix:RL}.
To simulate the continuous changes in upper body movements when controlled by our cockpit, we resample target upper-body poses every 1 second according to the above process. We then use uniform interpolation to ensure that the target movement gradually changes from the current value to the desired value over the 1-second interval. Without this approach, we find that the robot struggles to maintain balance under continuous motions.

\begin{figure}[!ht]
  \centering
  \includegraphics[width=0.48\textwidth]{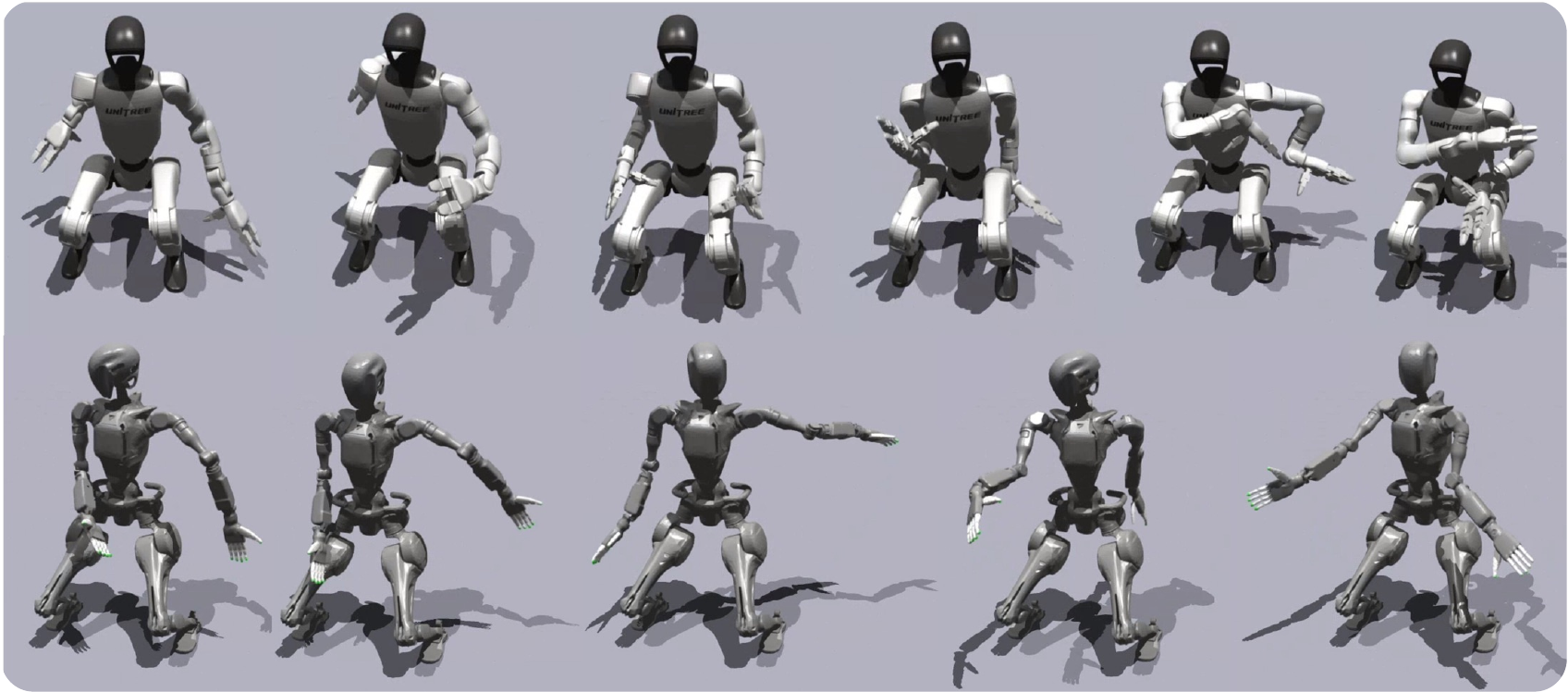}
  \caption{Different robots are trained to walk and squat with continuous changing upper-body poses in Isaac Gym.}
  \label{fig:gym}
\end{figure}
\vspace{-2pt}
\subsubsection{Height Tracking Reward}
\label{sec:height}
Tracking heights can significantly expands the feasible operational workspace of humanoid robots, thus helping the robots perform more loco-manipulation tasks. Therefore, $\pi_{loco}$ needs to enable the robot to squat to the target height $h_t$. To achieve this, we design a new reward function
\begin{equation}
\label{eq:rknee}
    r_{knee} = -\|(h_{r,t}-h_{t})\times(\frac{q_{knee,t}-q_{knee,min}}{q_{knee,max}-q_{knee,min}}-\frac{1}{2})\|,
\end{equation}
where $h_{r,t}$ is the robot's actual height, $q_{knee,min}$ and $q_{knee,max}$ are the maximum and minimum actions of knee joints, $q_{knee,t}$ is the current positions of robot's knee joints. $r_{knee}$ encourages flexion of the knee joints when $h_{r,t}< h_t$, and encourages extension when $h_{r,t}> h_t$. In the training process, we resample all commands every 4 seconds. At this point, one-third of the environments are randomly selected to train the robot to squat, while the remaining two-thirds focus on teaching the robot to stand and walk. This strategy helps balance the learning of squatting and walking. Additionally, the same environment switches between learning to squat and learning to walk, enabling the policy to smoothly transition between squatting and walking tasks. For better understanding of \cref{eq:rknee}, we visualize it in Appendix \ref{appendix:RL}.

\subsubsection{Symmetry Utilization}
We introduce the same trick as \cite{su2024leveraging} to our training framework. Each time we obtain a transition $T_t=(s_t, a_t, r_t, s_{t+1})$ from the simulation, we perform a flip operation on it. Specifically, we apply symmetry to the actor and critic observations with respect to the robot's x-z plane. This involves flipping elements such as the positions, velocities, and actions of the robot's left and right joints, as well as the desired turning velocity, across the x-z plane to obtain a mirrored transition $T_t'$. Both $T_t$ and $T_t'$ are then added to the rollout storage. This process helps to improve data efficiency and ensure symmetry in the sampled data, reducing the likelihood of the trained policy being asymmetrical in terms of left and right performance. In the learning phase, we also apply this procedure to the samples $T_t$ got from the rollout storage to get $T_t'$. Both $T_t$ and $T_t'$ are passed through the actor and critic networks to obtain $a_t$, $a_t'$, $V_t$, $V_t'$ respectively, which are used to calculate additional losses:
\begin{equation}
    \mathcal{L}^{actor}_{sym}=MSE(a_t,a_t'),
\end{equation}
\begin{equation}
    \mathcal{L}^{critic}_{sym}=MSE(V_t,V_t').
\end{equation}
These two losses are added to the network optimization process, thereby enforcing symmetry of the neural network.

\begin{figure*}[!ht]
  \centering
  \includegraphics[width=0.98\textwidth]{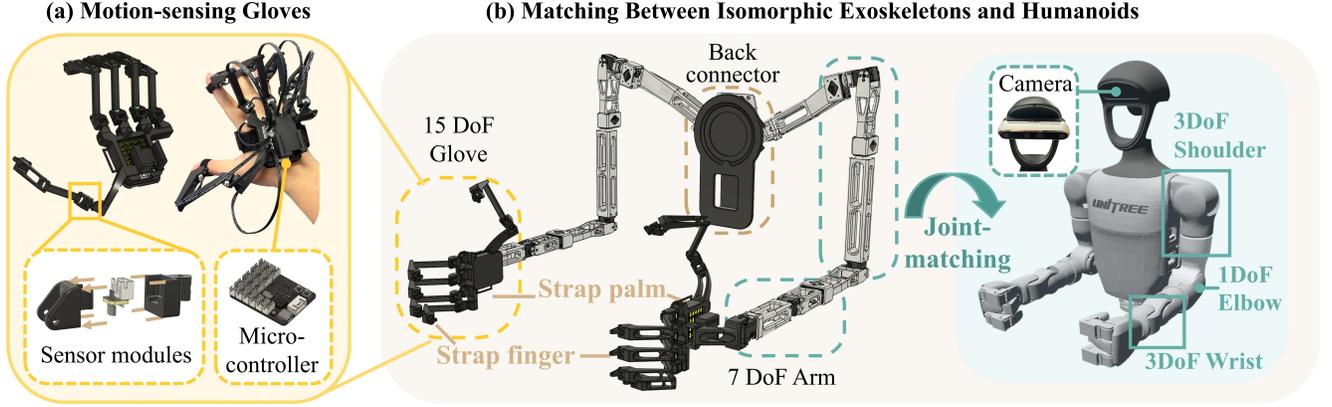}
  \caption{\textbf{Upper-body Exoskeleton.} \textcolor{mycolor}{(a):} The model architecture and physical demonstration of motion-sensing gloves. \textcolor{mycolor}{(b):} The structural architecture of the upper-body exoskeleton system, comprising an isomorphic exoskeleton and motion-sensing gloves, with kinematic mapping methodology between the isomorphic exoskeleton and Unitree G1.}
  \vspace{-10pt}
  \label{fig:Upper-body Exoskeleton}
\end{figure*}

\subsection{Hardware System Design}
\label{sec:hard_design}
To enable a single operator to control the full body of humanoid robots, we design a low-cost  exoskeleton-based hardware system as shown in the left part of \cref{fig:overview}. For the upper-body teleoperation of the humanoid robots, we  design 3D-printed 7-DoF isomorphic exoskeleton arms for precise mapping of the upper limb joint angles, specifically tailored for two types of humanoid robots: Unitree G1 and Fourier GR-1. Additionally, we design a pair of low-cost motion-sensing gloves capable of mapping up to  to 15 DoF of finger angles. For locomotion command acquisition, we design a foot pedal that simulates the press-and-release actions of the foot during driving, enabling control of the humanoid robot's movements such as walking and squatting. The operator can easily deploy this system and perform single-person teleoperation of the robot's loco-manipulation, similar to driving a car in a cockpit or playing a racing game.

\subsubsection{\textbf{Isomorphic Exoskeleton}}
To achieve accurate control and mapping of the upper limb joints of the humanoid robots, we employ an isomorphic exoskeleton as the teleoperation solution for controlling the robot's upper body. Based on the morphology of the Unitree G1 and Fourier GR-1, our isomorphic exoskeleton design consists of a symmetric pair of arms, each with 7 DoF, corresponding to the 7 DoF of each arm of the robot (3 DoF for the shoulder, 1 DoF for the elbow, and 3 DoF for the wrist), \red{as shown in \cref{fig:Upper-body Exoskeleton}}\textcolor{mycolor}{(b)}. Each joint of the exoskeleton is equipped with a DYNAMIXEL XL330-M288-T servo, which provides joint angle readings and adjustments with an accuracy of 0.09°, enabling precise joint angle mapping and initial calibration. \red{Compared to incremental encoders, servo motors can store absolute position and retain the current position data even after power loss, thus eliminating recalibration upon restarts.} The exoskeleton's operational part is designed to match the length of the human arms. Considering the challenge of fully replicating the robot's upper arm structure, we align the servos with the robot's motor URDF joint coordinate system. \red{We can obtain the offsets $o_t$ between the servos' position angles $p_t$ and the robot's joint angles $q_t$. Since Dynamixel servos can store absolute positions, $o_t$ remains fixed after assembly. Additionally, the servo disc has four symmetric holes, which causes $o_t$ to be an integer multiple of $\pi/2$. We use 
\begin{equation}
    q_t = \pm k_t(p_t+\frac{n_t\pi}{2})+\tau_t
\end{equation}
to achieve kinematic equivalence and calibration, where the offset follows $o_t =  \frac{n_t\pi}{2},\quad n_t \in \mathbb{Z}$ , with $\pm k_t$ is a coefficient that adjusts the direction and scale of the angle change, and $\tau_t$ is the additional joint angle compensation. We set $k_t = 1$ and $\tau_t = 0$, meaning no additional scaling or angle compensation is needed.}

\subsubsection{\textbf{Motion-sensing Gloves}}
For fine teleoperation of the fingers, we adopt a joint-matching approach. Based on the Nepyone glove project~\cite{Nepyope2023Project-Homunculus}, we design a low-cost motion-sensing glove that connects directly to the exoskeleton for assembly and use, providing up to 15 DoF for finger capture to control dexterous hands. Specifically, each finger is equipped with three sets of sensors, which map the pitch motion of the finger tip and finger pad, as well as the yaw motion of the finger pad. This setup is sufficient to enable the mapping of different dexterous hands for humanoid robots. We place Hall effect sensors and small neodymium magnets at each joint. When the joint rotates, the neodymium magnet rotates as well, thereby affecting the magnetic field sensed by the sensor and achieving the mapping of finger joint angles. Additionally, we design the glove's microcontroller, sensor modules, and structural model. The microcontroller is mounted on the back of the hand and can be directly connected to the packaged sensors using terminal connectors, allowing for easy plugging and unplugging to reassign and modify the mapping relationships, as shown in \red{as shown in \cref{fig:Upper-body Exoskeleton}}\textcolor{mycolor}{(a)}. Our motion-sensing gloves can be easily attached to and detached from different exoskeletons, offering high versatility.

\begin{figure}[!ht]
  \centering
  \includegraphics[width=0.42\textwidth]{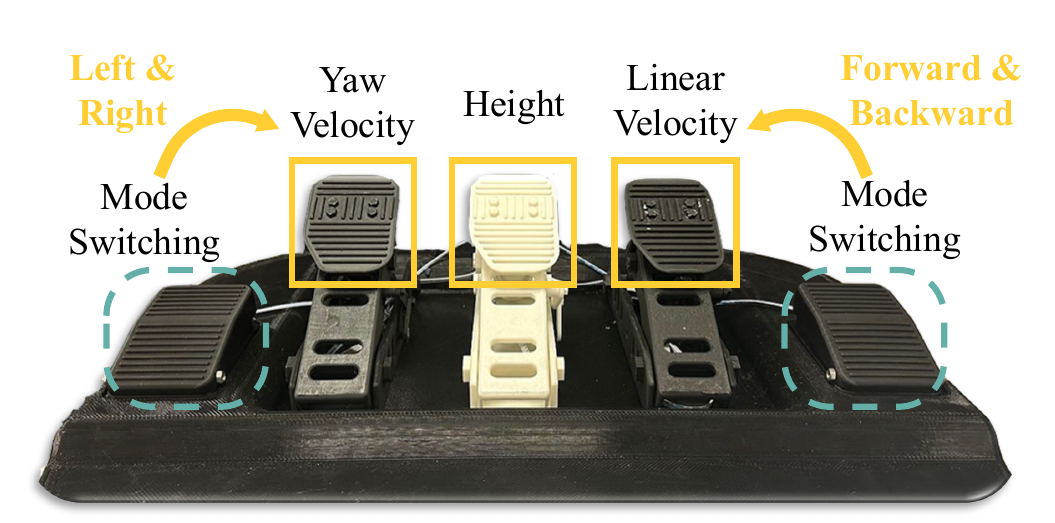}
  \caption{\textbf{Pedal command control.} The three small pedals respectively control $ [0, \pm{\omega_{max}}] $, $ [{H_{min}}, {H_{max}}] $, and $ [0, \pm{V_{max}}] $. The left-side switching button is used to toggle between left and right modes, while the right-side switching button is used to toggle between forward and backward modes.}
  \vspace{-5pt}
  \label{fig:pedal}
\end{figure}

\subsubsection{\textbf{Foot Pedal}}
\begin{figure*}[!ht]
  \centering
  \includegraphics[width=1.00\textwidth]{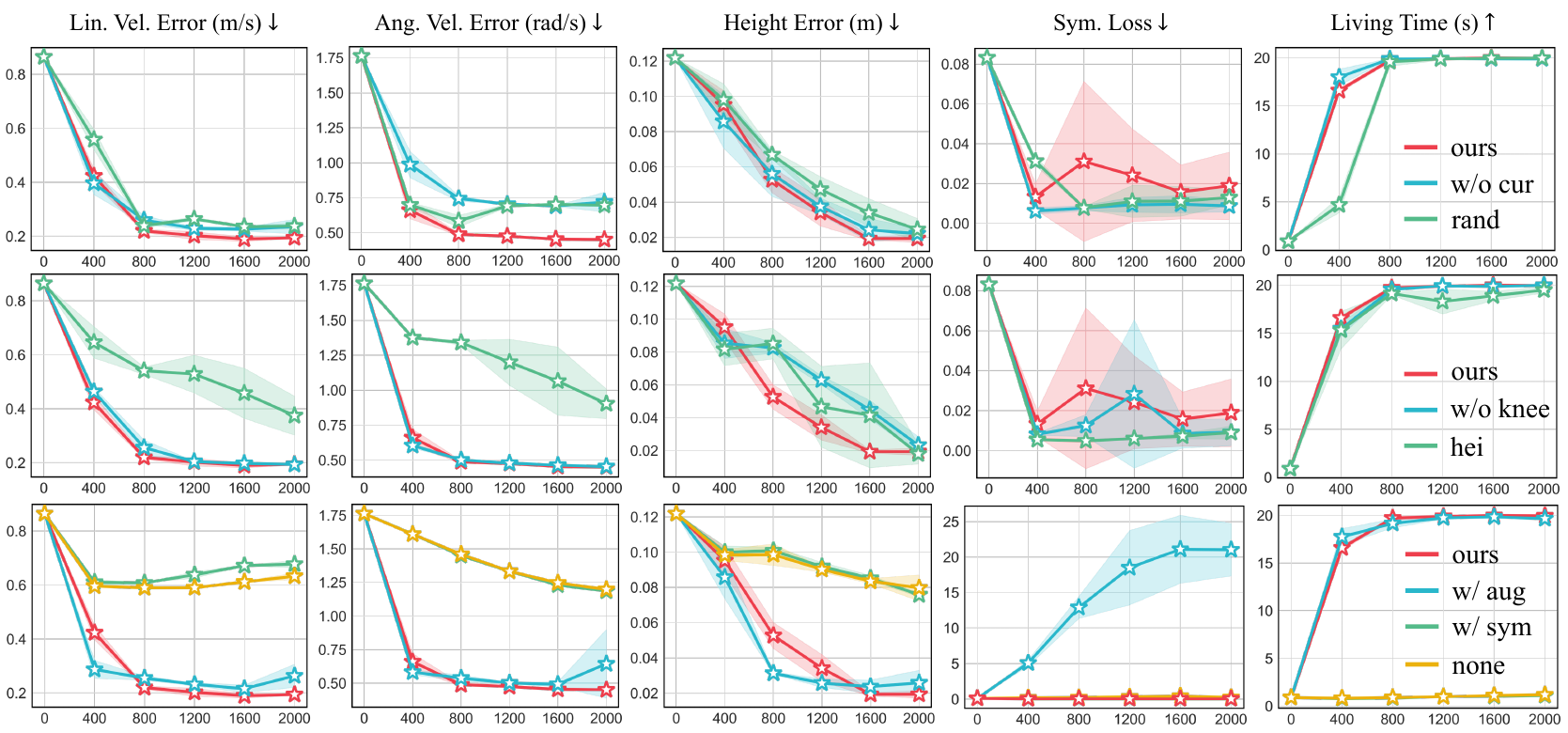}
  \caption{\textbf{Ablation experiments of our RL training framework.} Each row from top to bottom represents the ablation study for upper-body curriculum, height tracking reward, and symmetry utilization, respectively. Each column represents the evaluation of the corresponding metrics for checkpoints under different ablation settings. The $\uparrow$ and $\downarrow$ symbols beside the metrics indicate whether a higher or lower value is better for the respective metric.}
  \label{fig:rl_exp}
  \vspace{-8pt}
\end{figure*}
In our cockpit, the foot pedal is used as a replacement for a remote controller, enabling command control of the humanoid robot's lower body by giving commands $C_t$ to $\pi_{loco}$. The operator controls the acceleration and deceleration of the robot's lower body movement by pressing and releasing
the foot pedal.
We use high-precision rotary potentiometers to map pedal pressure changes to electrical signals. In our system, we control the humanoid robot's locomotion through linear velocity, yaw velocity, and height adjustment. These commands allow  the robot to fully demonstrate its locomotion capabilities. To achieve this, we use three small pedals to control these commands. Additionally, a pair of mode-switching buttons (foot-operated with momentary switches) are used to toggle between forward/backward and left/right turning directions, as shown in \cref{fig:pedal}. Users can modify the pedal configuration and reassign commands to adapt to diverse movement combinations.

\section{Experiments}

\subsection{\red{Humanoid Whole-body Control}}

\subsubsection{Ablation of training framework}
\label{sec:abla}
In this section, we perform ablation experiments on the proposed upper-body pose curriculum, the height tracking reward, and the use of symmetry. All ablation experiments are conducted based on the methods described in \cref{sec:method_rl}. For each setting, we use three random seeds to train policies for Unitree G1 and evaluate them in 1000 environments over a 20-second evaluation period with random upper-body poses sampled from \cref{equa:sample} with $\red{\rho_a}\rightarrow1$. Metrics for evaluation are tracking linear velocity error, tracking angular velocity error, tracking height error, symmetry loss and living time. The final performance for each setting is obtained by computing the average and standard deviation of the results across the three policies trained from three random seeds. All trainings are conducted on Nvidia RTX 4090 and simulated by Isaac Gym with 4096 parallel environments, where components unrelated to the ablation are kept unchanged, and only relevant parts are modified for training. Detailed parameters used in training and evaluation processes are listed in Appendix \ref{appendix:RL}. We mark the setting of our proposed method as \textbf{ours} in the following sections.

\textbf{Upper-body Pose Curriculum.} We compare \textbf{ours} against two alternatives: \textbf{w/o cur}, which omits the curriculum and directly samples $a_i=\mathcal{U}(0,\mathcal{U}(0,1))$, and \textbf{rand}, which uses the same $\red{\rho_a}$ curriculum but replaces \cref{equa:sample} with $a_i=\mathcal{U}(0,\mathcal{U}(0,\red{\rho_a}))$. Since all three methods adopt the same sampling strategy $a_i=\mathcal{U}(0,\mathcal{U}(0,1))$ as $\red{\rho_a}\rightarrow1$, the final objective remains consistent, ensuring a fair comparison. The experimental results, shown in the first row of \cref{fig:rl_exp}, reveal that \textbf{ours} outperforms both \textbf{w/o cur} and \textbf{rand} in linear velocity tracking, angular velocity tracking, and height error, with faster convergence and smaller errors. There is no significant difference between \textbf{w/o cur} and \textbf{rand} in the final results for these metrics. Given that the symmetry loss can reach values on the order of 20 without constraints, no significant difference is observed across the three methods in terms of symmetry loss. All three configurations achieve similar final living times, but \textbf{ours} and \textbf{w/o cur} converge more quickly. \textbf{Rand}, despite employing some curriculum adjustments, is limited by $\red{\rho_a}$, and values in the range $(\red{\rho_a},1]$ are not sampled during training, making it harder for the model to converge as $\red{\rho_a}$ increases. In contrast, both \textbf{ours} and \textbf{w/o cur} sample the full $[0,1]$ range from the beginning, enabling faster and more stable convergence. Thus, our curriculum approach leads to better performance compared to \textbf{rand}. Although \textbf{w/o cur} does not use a curriculum, allowing $a_i$ to continuously sample from $[0,1]$, the lack of difficulty smoothing leads to worse final tracking results, highlighting that our curriculum design offers a more effective training process.

\textbf{Height Tracking Reward.} We design two additional algorithms \textbf{w/o knee}, which does not use $r_{knee}$ described in \cref{eq:rknee} and \textbf{hei}, which also omits $r_{knee}$ but increases the scale of the height tracking reward. We show the results in the second row of \cref{fig:rl_exp}. As shown in the figure, none of the three settings cause significant changes in the symmetry loss during training. In terms of linear velocity error and angular velocity error, \textbf{ours} and \textbf{w/o knee} perform similarly, while \textbf{hei} shows much larger errors. For height error, our method converges faster than both \textbf{w/o knee} and \textbf{hei}, even though \textbf{hei} initially performs better (at 400 steps). There is no significant difference among the three settings in terms of living time. These results indicate that just scaling up the height tracking reward in \textbf{hei} may initially lead to faster reduction in height tracking error, but it negatively affects the feedback from other rewards, preventing the robot from balancing multiple tasks effectively. In fact, \textbf{hei} ultimately does not achieve faster convergence in height tracking compared to \textbf{ours}. In contrast, the inclusion of \textbf{rknee} in our method provides more specific guidance for squat tracking, allowing the robot to reduce tracking error and converge more quickly. This highlights the effectiveness of \textbf{rknee} in helping the robot learn squat motions. 

\textbf{Symmetry Utilization.} We introduce three algorithmic variants for comparison with \textbf{ours} in terms of symmetry utilization:  \textbf{w/ aug}, which uses only symmetrical data augmentation; \textbf{w/ sym}, which only uses symmetry loss; and \textbf{none}, which does not employ symmetrical data augmentation or symmetry loss. Testing results are presented in the third row of \cref{fig:rl_exp}. Except for symmetry loss, the performance of \textbf{ours} and \textbf{w/ aug} is similar. However, when considering overall tracking accuracy, ours performs slightly better. On the other hand, \textbf{w/ aug} exhibits a very high symmetry loss, suggesting that using symmetry loss helps maintain the robot's left-right symmetry in the learned policy. This indirectly supports the idea that a symmetric policy benefits the robot’s locomotion tasks~\cite{su2024leveraging}. Both \textbf{nsym} and \textbf{none} show a tendency for improvement, but their training speed is much slower. Notably, a direct comparison between \textbf{w/ sym} and \textbf{none} reveals that \textbf{w/ sym} achieves lower symmetry loss. However, due to slower training, \textbf{none} exhibits less symmetry breaking compared to \textbf{w/ aug}. In summary, symmetry data augmentation significantly improves training efficiency, while the use of symmetry loss effectively prevents the policy from sacrificing symmetry to complete tasks and also benefits the task itself.

\begin{figure}[!ht]
\setlength{\belowcaptionskip}{-10pt}
  \centering
  \includegraphics[width=0.35\textwidth]{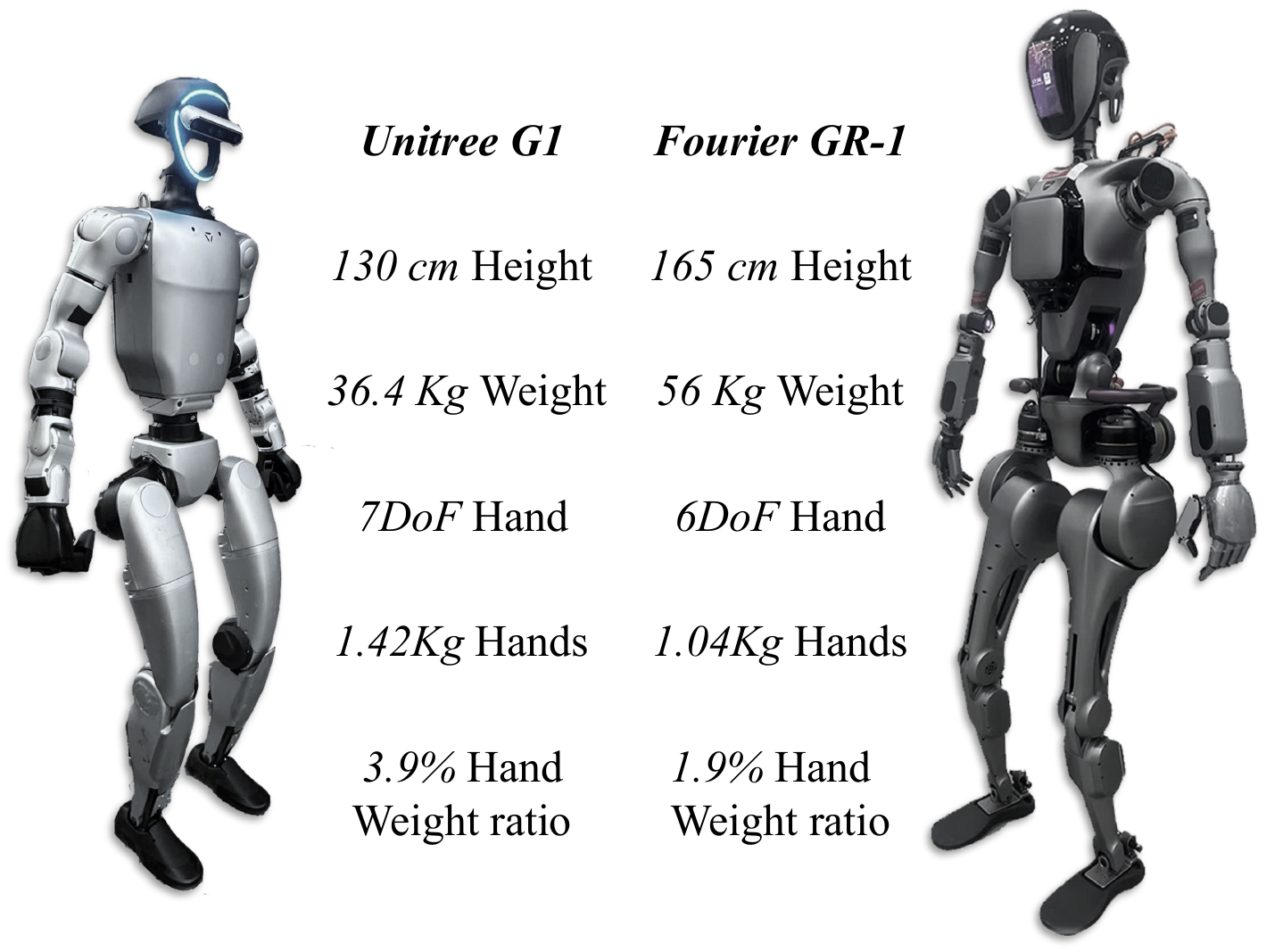}
  \caption{Key parameters of Unitree G1 (left) and Fourier GR-1 (right). Hand weight ratio = total weight / hands weight.}
  \label{fig:g1 & gr1}
\end{figure}
\subsubsection{Training on Different Robots}

We select another kind of robot, Fourier GR-1, which is quite different from Unitree G1, to demonstrate the generality of our approach across different robot models. As shown in \cref{fig:g1 & gr1}, Fourier GR-1 is much taller and heavier than Unitree G1 while having lower hand weight ratio. Compared to the training setting of Unitree G1, we only change the range of height tracking and some robot-specific distance values, without any other changes in reward scales or training pipeline. We evaluate the policy trained after 2k steps of each robot with metrics used in \cref{sec:abla} and present them in \cref{tab:gr1}. The results demonstrate that even though these two kinds of robots are quite different, our RL training framework can train them to converge to a policy which can drive robots to perform locomotion and squatting tasks robustly under any upper-body poses. Training details for Fourier GR-1 can be found in Appendix \ref{appendix:RL}.

\begin{table}[!ht]
    \centering
    \caption{Evaluation of different robots trained with our RL training framework}
    \resizebox{.47\textwidth}{!}{
    \begin{tabular}{l|cc}
        \toprule
        \textbf {Metrics} & Unitree G1 & Fourier GR-1 \\
        \midrule
        Lin. Vel Error (m/s)  & 0.194\ci{0.003} & 0.273\ci{0.003}\\
        Ang. Vel Error (rad/s) &  0.451\ci{0.006} & 0.540\ci{0.002}\\
        Height Error (m) & 0.022\ci{0.019}& 0.038\ci{0.003}\\
        symmetry loss (-) &  0.019\ci{0.017}& 0.009\ci{0.001}\\
        Living Time (s) &  19.947\ci{0.092} & 19.960\ci{0.035}\\
        \bottomrule
    \end{tabular}}
    \label{tab:gr1}
\end{table}


\subsection{Teleoperation Hardware Performance}
We list a series of hardware indicators for our teleoperation hardware system consisting of isomorphic exoskeleton arms, a pair of motion-sensing exoskeleton gloves, and a pedal in \cref{tab:Hardware}. We detail their costs, with the primary expense attributed to the exoskeleton section. This is because we independently design and solder the control boards (PCBs) and sensor modules for the motion-sensing gloves and the pedal components. The acquisition frequency represents the update signal frequency measured between the hardware components of the teleoperation system and the host computer via a wired connection at a baud rate of 115200. Changing the baud rate can affect the acquisition frequency. The acquisition accuracy represents the range of angular change (in degrees) and the corresponding variation in acquisition readings, ranging from 0 to 4095(~$2^{12}$). Since the mapping relationship for the motion-sensing gloves is not a clearly defined linear one, and the mapping angles for each finger joint vary, more detailed information can be found in the Appendix~\ref{sec:Glovesc}.
\begin{table}[!ht]
    \centering
    \caption{Hardware Indicators of three component of the hardware system.(Freq.:frequency, Acc.:accuracy)}
    \resizebox{.47\textwidth}{!}{
    \begin{tabular}{lccc}
       \toprule%
        Hardware&Cost (\$) &Acquisition Freq.&Acquisition Acc. \\
        \cmidrule[0.6pt](lr{0.125em}){1-4}
        \textbf {Exoskeleton}& 430 &0.26 kHz&~$2^{12}$(with 360\textdegree)\\
        \textbf {Glove}& 30 (each)&0.3 kHz& ~$2^{12}$ \\
        \textbf {Pedal}& 20 &0.5 kHz& ~$2^{12}$(with 270\textdegree)\\
       \bottomrule
    \end{tabular}}
    \label{tab:Hardware}
\end{table}
For upper-body teleoperation, the task can be divided into two parts: arm control and dexterous hand control. We select the arm pose frequency and hand pose frequency as evaluation metrics, which measures the smoothness and fluidity of teleoperation. In \cref{tab:Fre.}, we compare the visual and VR schemes with our joint-matching scheme. Since our joint-matching scheme directly sets the robot's upper-body poses without the need for additional time-consuming processes, the output frequency to the robot closely matches the acquisition frequency. Therefore, our approach achieves a very high output frequency without requiring GPU and System on Chip (SoC) intensive hardware.
\begin{table}[!ht]
    \centering
    \caption{Upper-body teleoperation frequency of output to the robot's arm and hand. (SoC: System on Chip)}
    \resizebox{.47\textwidth}{!}{
    \begin{tabular}{l|ccc}
        \toprule
        \textbf {Teleop system} & Hardware & Arm (Hz) & Hand (Hz) \\
        \midrule
        Telekinesis~\cite{Sivakumar-RSS-22}  & 2 RTX 3080 Ti & 16 & 24\\
        AnyTeleop~\cite{qin2023anyteleop} &  RTX 3090 & 125 & 111\\
        OpenTeleVision~\cite{cheng2024open} & M2 Chip & 60 & 60\\
        \ourrow Ours & \textbf{No GPU / SoC} & \textbf{263}& \textbf{293}\\
        \bottomrule
    \end{tabular}}
    \label{tab:Fre.}
\end{table}

\begin{figure}[!ht]
  \centering
  \includegraphics[width=0.48\textwidth]{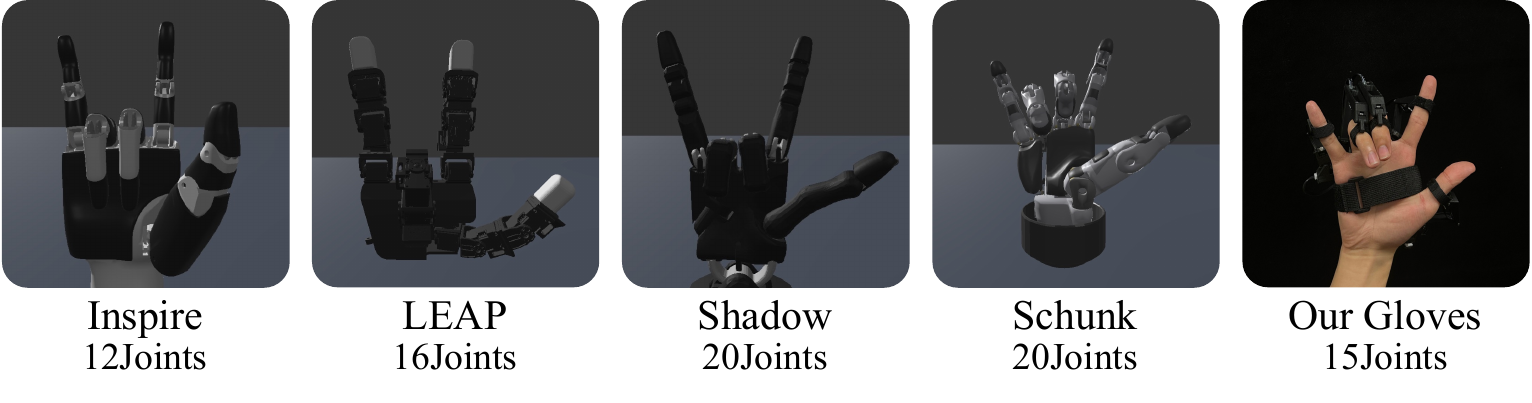}
  \caption{Controlling different types of dexterous hands in simulation with our motion-sensing gloves.}
  \label{fig:hand}
  \vspace{-5pt}
\end{figure}

To further demonstrate the extensibility of our motion-sensing gloves, we test different types of dexterous hands from the Dex Retargeting library in AnyTeleop~\cite{qin2023anyteleop} within the SAPIEN~\cite{Xiang_2020_SAPIEN} environment. The results are presented in \cref{fig:hand}, with the upper line shows names of tested dexterous hands while the lower line indicates number of joints of each hand.

\subsection{Teleoperation System}
\subsubsection{Real World}
\begin{figure}[!ht]
  \centering
  \includegraphics[width=0.48\textwidth]{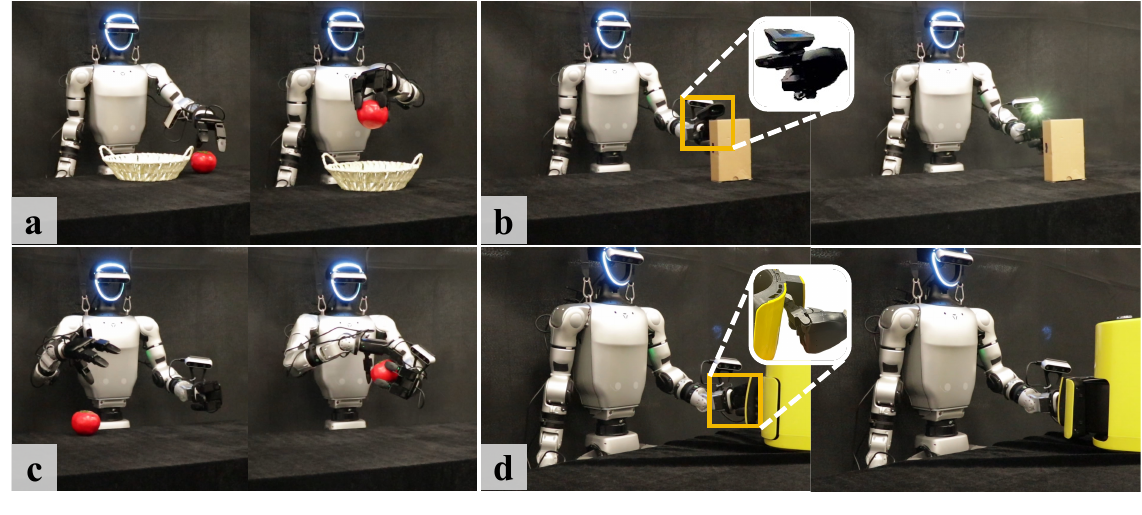}
  \caption{Desktop tasks for comparison of completion time. \textcolor{mycolor}{a:} Pick \& Place; \textcolor{mycolor}{b:} Scan Barcode; \textcolor{mycolor}{c:} Hand Over; \textcolor{mycolor}{d:} Open Oven.}
  \vspace{-5pt}
  \label{fig:compare}
\end{figure}
We deploy the trained policy on the Unitree G1 in the real world and teleoperate it to perform various loco-manipulation tasks using our isomorphic exoskeleton hardware system. \red{We employ WiFi for communication between the cockpit and the robot. Since our system requires only 128 bytes(32-bit floats) per data packet, the measured communication latency under normal network conditions is 16 ms — a result considered acceptable for real-time control. Furthermore, we implement network techniques including checksum verification to guarantee data transmission integrity.} The deployment code for G1 is derived from~\cite{margolis2023walk}. \cref{fig:teaser} \textcolor{mycolor}{(a)} and \cref{fig:teaser} \textcolor{mycolor}{(c)} demonstrate the robot's capability to squat, pick objects from lower shelves, and place them on higher ones, as well as to grasp and transfer boxes between shelves utilizing its locomotion abilities. \cref{fig:teaser} \textcolor{mycolor}{(b)} highlights the extensibility of our system, enabling two operators to control separate robots and collaboratively perform tasks, such as transferring apples. In \cref{fig:teaser} \textcolor{mycolor}{(d)}, the robot is controlled to push a 60 kg person sitting in a chair, who weighs roughly twice as much as the robot, demonstrating the robustness of the loco-manipulation system. \cref{fig:teaser} \textcolor{mycolor}{(e)} illustrates how the robot uses its loco-manipulation abilities to open an oven by grasping the handle and moving backward simultaneously. \cref{fig:teaser} \textcolor{mycolor}{(f)} shows that our teleoperation system is capable of performing dual-hand collaborative tasks, such as one hand passing an object to the other. \cref{fig:teaser} \textcolor{mycolor}{(g)} demonstrates the robot's ability to grasp objects from low ground, while \cref{fig:teaser} \textcolor{mycolor}{(h)} shows the robot’s capability to lift and place heavy items, such as a bundle of flowers, into a box using both arms. \cref{fig:teaser} \textcolor{mycolor}{(i)} demonstrates how the robot maintains balance with different upper-body poses. In all these tasks, each robot is controlled by a single operator, and the communication between the robot and operator is facilitated via Wi-Fi, without restricting the robot's movement space. These tasks showcase the robustness of our loco-manipulation policy and \ourshort’s ability to teleopeate humanoids perform a wide range of complex tasks in various environments.
\begin{figure}[!ht]
  \centering
  \includegraphics[width=0.46\textwidth]{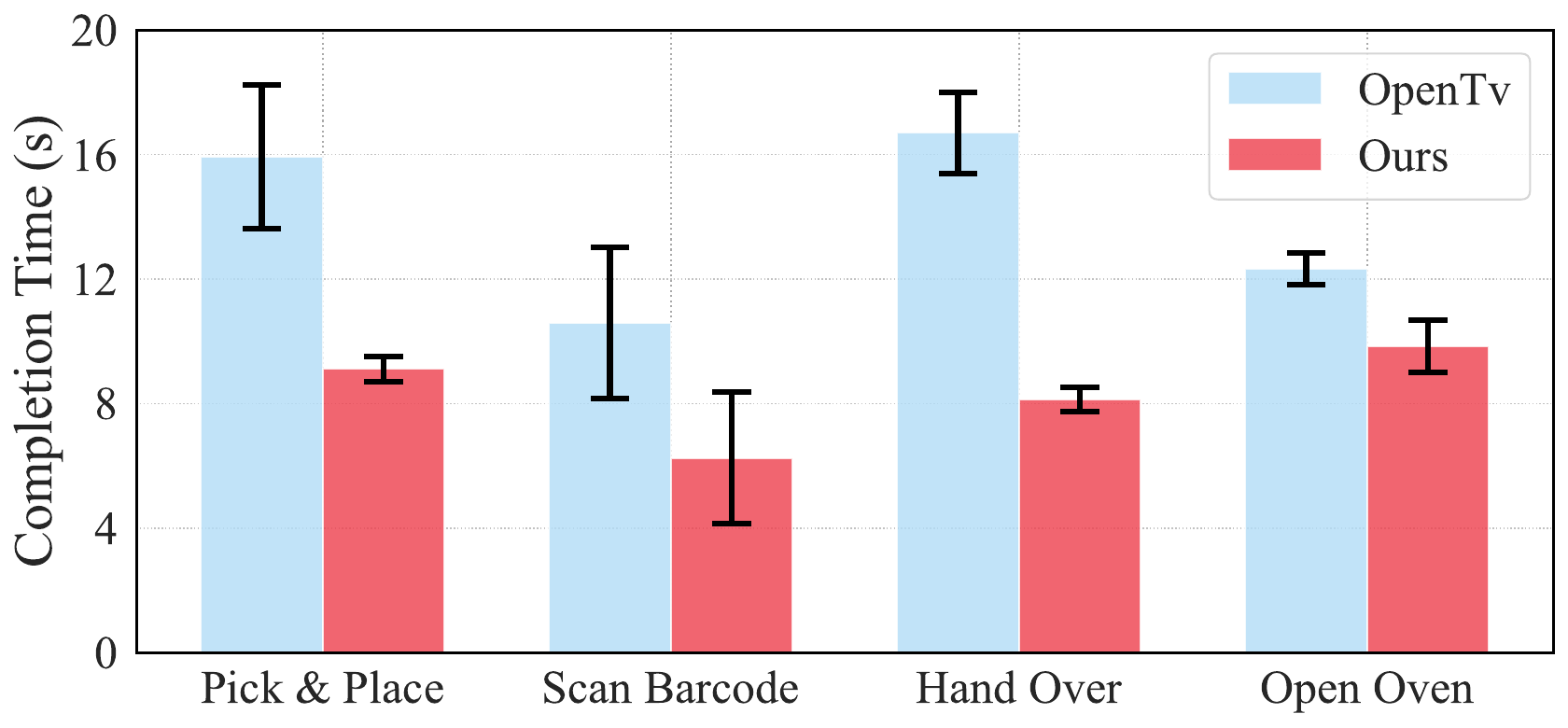}
  \caption{Comparison of completion time     to perform desktop tasks between our hardware system and OpenTelevision~\cite{cheng2024open}.}
  \label{fig:opentv}
\end{figure}

To demonstrate the efficiency of our teleoperation system, we compare the task completion time between our hardware system and a VR-based method, OpenTelevision~\cite{cheng2024open}, across four tasks as shown in 
\cref{fig:compare}. These tasks are designed to evaluate the system's ability to precisely control the robot's arms and hands in various 
scenarios: \textbf{Pick \& Place:} The robot is required to grasp a tomato from the table and place it into a fruit basket. \textbf{Scan Barcode:} 
The robot must hold a scanner, press its button using a finger, and scan a barcode on a box. \textbf{Hand Over:} The robot needs to grasp a tomato and pass it to another hand. \textbf{Open Oven:} The robot must insert its finger into a handle and open the oven door. These tasks test key 
capabilities of teleoperation, including precise positioning, bimanual coordination, and fine-grained finger control. The results, shown in 
\cref{fig:opentv}, indicate that our system achieves task completion times nearly half of those of the VR-based method. Notably, when tasks require precise positioning and orientation, the performance gap between our system and the VR method becomes even more pronounced. This is because VR-based pose estimation tends to perform poorly in tangential directions, whereas our exoskeleton-based approach avoids such issues entirely. These results demonstrate that our exoskeleton system enables operators to 
teleoperate robots more smoothly and efficiently, particularly in tasks requiring high precision and dexterity.

\subsubsection{\red{User Study}}
\begin{figure}[!ht]
  \centering
  \includegraphics[width=0.4\textwidth]{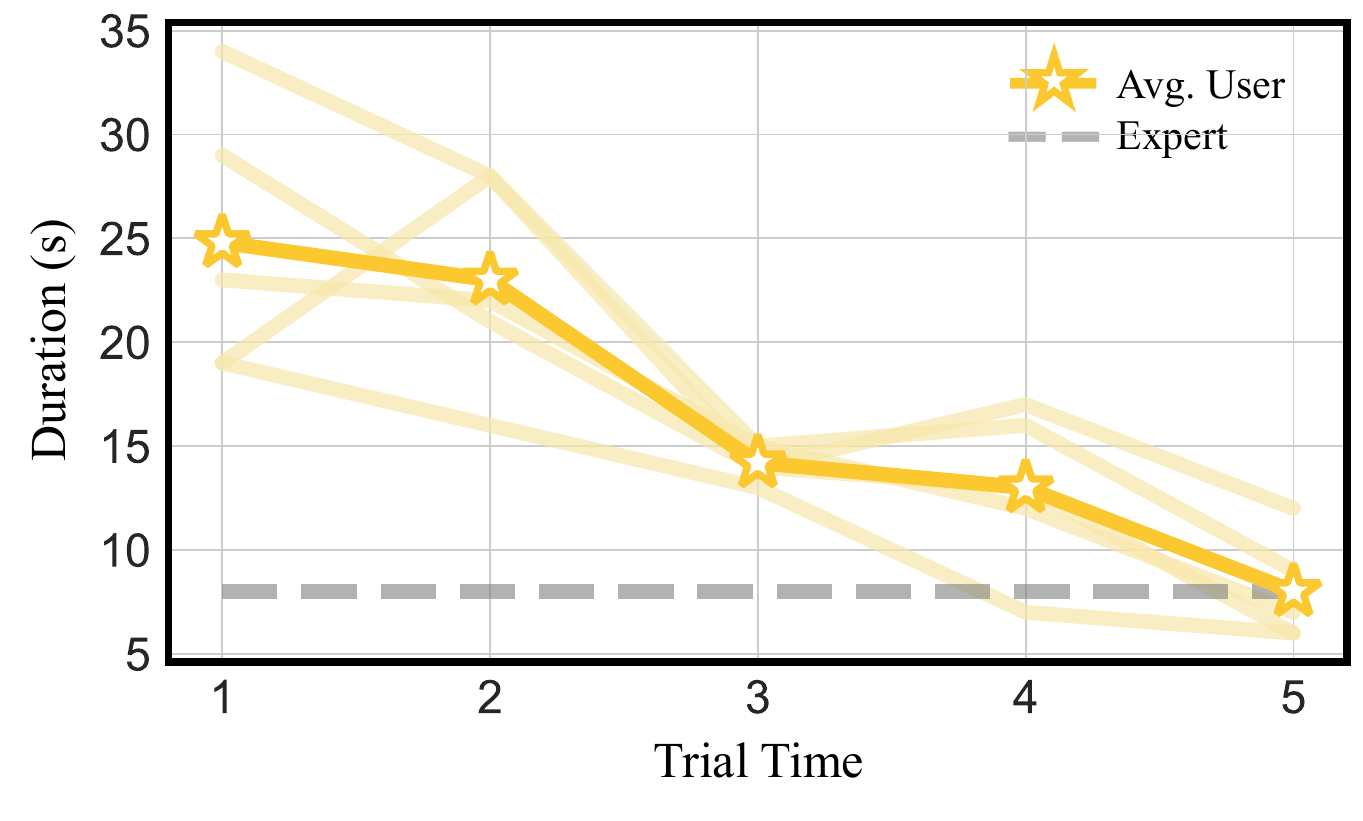}
  \vspace{-7pt}
  \caption{\red{\textbf{Teleoperation learning curves\cite{huang20243dvitac}.} New users can quickly achieve expert speed on a new task with our hardware.}}
  \label{fig:user}
\end{figure}
\red{We recruit five testers with varying heights, weights, and genders to evaluate our hardware system for teleoperating Unitree G1 in \textbf{Hand Over} task. All novices don't have any prior experience using the system or teleoperating humanoid robots. Each trial begins with the operator positioning both the robot and a tomato at identical designated starting points. Upon the "Start" command, timing commences as the operator guides the robot to first grasp the tomato with one manipulator, then transfer it to the secondary manipulator. The trial concludes when the tomato is securely gripped by the receiving manipulator and the initial manipulator has fully retracted. The elapsed time from initiation to successful transfer completion is recorded as the "completion time," serving as the primary metric for evaluating operator proficiency with this system. After a brief tutorial, we record their completion times across five consecutive attempts. As shown in~\cref{fig:user}, the average novice time progressively approached expert-level performance despite significant physical differences among operators. This rapid improvement demonstrates our system's intuitive usability. This result also shows out system's strong adaptability to diverse body types with straps. Tester profiles with raw timing data are detailed in Appendix~\ref{sec:raw_data}, and the operational protocol in Appendix~\ref{sec:teaching}.}

\subsubsection{Simulation}
\begin{figure}[!ht]
  \centering
  \includegraphics[width=0.5\textwidth]{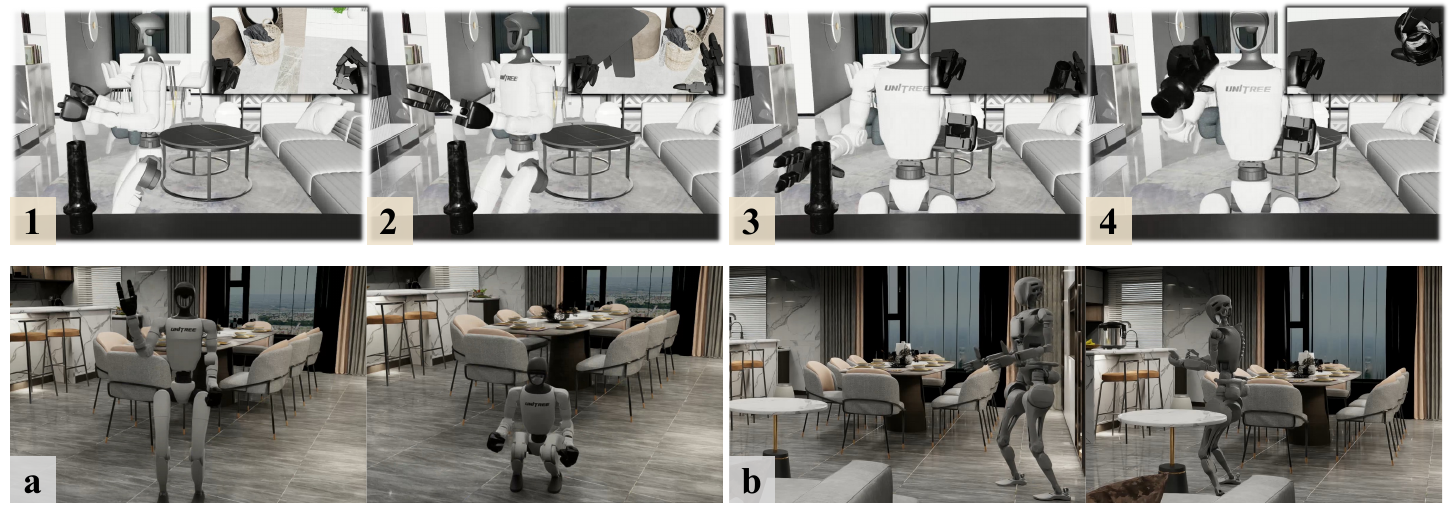}
  \caption{\textbf{Simulation migration.} The upper row illustrates how the operator controls the robot with FPV to perform loco-manipulation tasks. The lower row demonstrates the robot navigating through realistic simulated environments.}
  \vspace{-10pt}
  \label{fig:grutopia}
  
\end{figure}

We transfer the trained policies for Unitree G1 and Fourier GR-1 from Isaac Gym to a scene developed by GRUtopia~\cite{wang2024grutopia}, which is based on Isaac Sim and IsaacLab~\cite{mittal2023orbit}. This migration enables the use of~\ourshort to control robots within a variety of simulated environments. By leveraging these simulated scenes, the robots can perform diverse loco-manipulation tasks more cost-effectively and in a wider range of scenarios than would be feasible in the real world. As shown in \cref{fig:grutopia}, operators can seamlessly direct the robots' movements and actions in complex, realistic settings, demonstrating the versatility and applicability of \ourshort in diverse simulated contexts.
\subsection{Autonomous Policy}

\subsubsection{Data Collection}

\begin{wrapfigure}{l}{0.2\textwidth}
  \centering
  \includegraphics[width=0.18\textwidth]{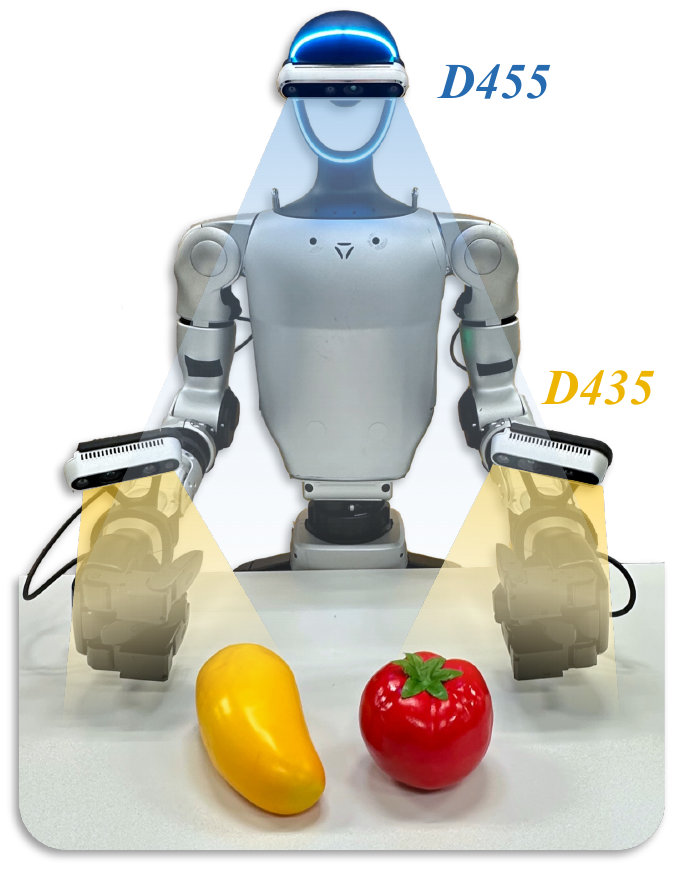}
  \caption{Hardware Setup for Imitation Learning}
  \label{fig:imi}
\end{wrapfigure}

To validate the effectiveness of the demonstratons collected by \ourshort for IL algorithms, we design two distinct tasks: \textbf{Squat Pick}: squatting to pick a tomato on the lower sofa; \textbf{Pick \& Place}: picking and placing a tomato. We capture RGB images, robot states $q_t$, the upper-body commands $q_{upper}$, and the locomotion commands $C_t$ at 10Hz, and collect 50 episodes per task. The hardware setup for image capture can be found in \cref{fig:imi}.

\subsubsection{Training Setting}
\label{sec:iltraining}
We adopt an end-to-end visuomotor control policy that takes images and robot proprioceptive signals as inputs and continuously outputs robot control actions.
We employ a model named Seer~\cite{tian2024predictive}, which features an autoregressive transformer architecture.
Multi-view images are processed through a MAE-pretrained ViT encoder, and the features of robot proprioceptive states are extracted using an MLP. These features are subsequently concatenated into tokens. The information of these tokens are then integrated by a transformer encoder. The transformer encoder utilizes an autoregressive method to generate latent codes for controlling upper arm joints, dexterous hand movements, and height commands. The final control action output is generated by three distinct regression heads. The whole network are optimized using SmoothL1 loss.
In real-world training scenarios, we configure the sequence length to 7, with both visual foresight and action prediction steps set to 3. We employ the MAE pre-trained ViT-B encoder, using bfloat16 configuration to speed up inference. This model is trained on eight A100 GPUs for 40 epoches, and we select the checkpoint with the lowest average validation loss for evaluation.
%
\begin{table}[!ht]
    \centering
    \caption{Success Rate of Imitation Learning Tasks}
    \begin{tabular}{c|cc}
        \toprule
        \textbf {Tasks} & Squat Pick & Pick \& Place\\
        \midrule
        Success Rate (\%)  & 73.3 & 80.0\\
        \bottomrule
    \end{tabular}
    \label{tab:imi}
\end{table}
\subsubsection{Learning Results}

After training with collected data, we deploy the trained model to humanoid robot in the real world, with the trained $\pi_{auto}$ taking over operator to control the robot.
\begin{figure}[!ht]
  \centering
  \includegraphics[width=0.48\textwidth]{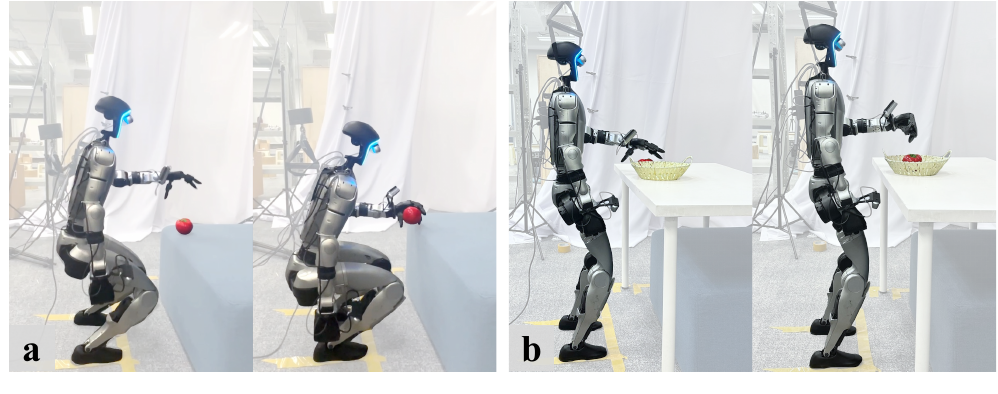}
  \caption{\textbf{Autonomous policy controlling robot to perform tasks.} \textcolor{mycolor}{a:} Squat Pick; \textcolor{mycolor}{b:} Pick \& Place.}
  \label{fig:imitask}
\end{figure}
We employ an Nvidia RTX 4080 to run the trained model and send the output to robot. The detailed deployment configuration are introduced in Appendix \ref{appendix:deploydetail}. For evaluation, we adopt the metric Success Rate (SR) of each task. After testing each proposed task for 15 times, we report the result as task success rate in \cref{tab:imi}. This result shows that data collected by our teleopertion system can actually drive robots to complete complex whole-body loco-manipulation tasks. Robots that controlled by $\pi_{auto}$ to perform proposed tasks are shown in \cref{fig:imitask}.

\section{Conclusion and Limitations}
In this paper, we introduce \ourshort, a novel humanoid teleoperation cockpit for humanoid loco-manipulation. With a low-cost isomorphic exoskeleton hardware system and a humanoid loco-manipulation policy trained by our RL training framework, ~\ourshort enables a single operator to teleoperate the whole body of humanoid robots and perform diverse loco-manipulation tasks either in the real world or in the simulation. Owing to the incorporation of an upper-body pose curriculum, a height-tracking reward, and symmetry-based techniques, Our training framework enables the development of robust loco-manipulation policies, ensuring stable walking and squatting capabilities across diverse robotic platforms, even under dynamically changing upper-body poses. Leveraging isomorphic exoskeleton arms, \ourshort enables significantly faster task execution than other systems, and our gloves are compatible with multiple kinds of dexterous hands. We present several ablation studies and real-world experiments to validate the robustness and accuracy of our system. In addition, we show the usability of collected data for IL.

\textbf{Limitations} Our policies still fall short of ensuring reliable traversal over diverse terrains. Additionally, the 15-DoF design of the motion-sensing gloves for the thumb does not fully align with human anatomy, resulting in less intuitive and smooth operation when controlling certain dexterous robotic hands. The current system lacks force feedback, which limits its effectiveness in applications requiring precise haptic interaction. \red{Furthermore, our exoskeleton doesn't afford teleoperation for waist even though our policy can support arbitrary upper-body poses.} Addressing these limitations will be a central focus of our future research efforts.

\section{acknowledgement}
This work is funded in part by the National Key R\&D Program of China (2022ZD0160201), and Shanghai Artificial Intelligence Laboratory. We would like to thank Huayi Wang, Shi Zhang, Zichao Ye, Hanqing Wang, Zirui Wang, Tao Huang, and Junli Ren for their valuable discussions. 


\bibliographystyle{unsrt}
\bibliography{references}

\clearpage
\newpage
\onecolumn
\begin{appendices}

\section{RL Training and Evaluation Details.}
\label{appendix:RL}
\subsection{Network Architecture}
\label{appendix:archi}
We use a network architecture similar to that employed in a previous quadruped locomotion work called HIM \cite{long2024hybrid}. The architecture consists of three main components: an estimator network $\mathcal{E}$, a follow-up network $\mathcal{N}$ that takes the output of $\mathcal{E}$ as part of its input, and a critic network $\mathcal{C}$. Together, $\mathcal{E}$ and $\mathcal{N}$ form the actor module, as illustrated in \cref{fig:rl}. All networks are implemented as 3-layer multilayer perceptrons (MLPs). Below, we describe their specific architectures, where $N_{joints} = N_{lower} + N_{upper}$, with $N_{lower}$ representing the number of lower-body joints and $N_{upper}$ the number of upper-body joints of the robot. Given that the dimensions of the input vectors are as follows: \red{$Dim([C_t, \omega_t, g_t]) = 3 + 3 + 3 = 9$ and $Dim([v_x, \omega_{yaw}]) = 2$}, we can determine the input dimensions for each network. The workflow is as follows: The sequence $O_{t-5:t}$ is fed into the encoder of $\mathcal{E}$. The encoder processes $O_t$ along with the ground truth values \red{$([v_x, \omega_{yaw}])$} and outputs the encoded information $\hat{I}_t$. $\hat{I}_t$ is then passed to $\mathcal{N}$ and is also used for contrastive learning with the output $\hat{I}'_t$ of the target network. Combining $\hat{I}_t$ and $O_t$, $\mathcal{N}$ produces the action $a_t$, whose dimension equals $N_{lower}$. The critic network $\mathcal{C}$ takes $O_t$ and the ground truth \red{$([v_x,  \omega_{yaw}])$} as its input.
\lstset{
    backgroundcolor=\color{gray!10},
    basicstyle=\ttfamily\small,
    breaklines=true,
    frame=single,
    keywordstyle=\color{blue},
    commentstyle=\color{green!50!black},
    stringstyle=\color{red},
    showstringspaces=false,
    mathescape=true
}
\begin{lstlisting}[language=Python, caption={Architecture of Neural Networks Used by Our RL Training Framework.}]
 ($\mathcal{E}$): HIMEstimator(
    (encoder): Sequential(
      (0): Linear(in_features=$6\times(9+2\times N_{joints}+N_{lower})$, out_features=256, bias=True)
      (1): ELU(alpha=1.0)
      (2): Linear(in_features=256, out_features=256, bias=True)
      (3): ELU(alpha=1.0)
      (4): Linear(in_features=256, out_features=35, bias=True)
    )
    (target): Sequential(
      (0): Linear(in_features=$9+2\times N_{joints}+N_{lower}$, out_features=256, bias=True)
      (1): ELU(alpha=1.0)
      (2): Linear(in_features=256, out_features=256, bias=True)
      (3): ELU(alpha=1.0)
      (4): Linear(in_features=256, out_features=32, bias=True)
    )
    (proto): Embedding(64, 32)
  )
  ($\mathcal{N}$): Sequential(
    (0): Linear(in_features=$35+9+2\times N_{joints}+N_{lower}$, out_features=512, bias=True)
    (1): ELU(alpha=1.0)
    (2): Linear(in_features=512, out_features=256, bias=True)
    (3): ELU(alpha=1.0)
    (4): Linear(in_features=256, out_features=256, bias=True)
    (5): ELU(alpha=1.0)
    (6): Linear(in_features=256, out_features=$N_{lower}$, bias=True)
  )
  ($\mathcal{C}$): Sequential(
    (0): Linear(in_features=$2+9+2\times N_{joints}+N_{lower}$, out_features=512, bias=True)
    (1): ELU(alpha=1.0)
    (2): Linear(in_features=512, out_features=256, bias=True)
    (3): ELU(alpha=1.0)
    (4): Linear(in_features=256, out_features=256, bias=True)
    (5): ELU(alpha=1.0)
    (6): Linear(in_features=256, out_features=1, bias=True)
  )
\end{lstlisting}

\subsection{Reward Scales}
\label{appendix:rwdscale}
We present the reward functions and their corresponding scales used in training the Unitree G1 and Fourier GR-1 robots in \cref{tab:reward}. Our reward functions adapt from PIM~\cite{long2024learninghumanoidlocomotionperceptive}, a previous work on humanoid locomotion, with several modifications and new additions. In addition to the "Squat knee" term $r_{knee}$ introduced in \cref{sec:height}, we revise the "Base height tracking" reward to align it with other tracking terms, as our objective is to track the changing target base height. Furthermore, we decompose the linear velocity tracking reward into separate components for x and y velocity tracking to better distinguish the robot's performance in these directions. To encourage the robot to remain stationary when zero velocity is required, we introduce a "stand still" reward. However, we observe that this reward can cause the robot to become overly inclined to maintain stillness, leading to instability during transitions between stationary and moving states. Experimental results indicate that reducing the $K_p$ of the ankle joint alleviates this issue. The rationale is that a lower $K_p$ tends to produce smaller torque outputs, making the ankle joint more responsive to positional changes when the center of gravity shifts. This positional change provides proprioceptive feedback to the policy, prompting necessary adjustments. The reward scales used in the training processes of these two heterogeneous robots are largely similar, further demonstrating the generality of our framework.

\begin{table}[!ht]
    \centering
    \caption{Reward Functions and Weights Used to Train Low-manipulation Policy}
    \renewcommand{\arraystretch}{1.5}
    \begin{tabular}{llcc} 
    \toprule[1.5pt] Reward & Equation & Weight of Unitree G1 & Weight of Fourier GR-1 \\ 
    \midrule[1.5pt] 
    x Vel. tracking & $\exp \left\{-4 \|\mathbf{v}_{x}-\mathbf{v}_{r,x}\|_2^2  \right\}$ & 1.5 & 1.5 \\
    
    y Vel. tracking & $\exp \left\{-4 \|\mathbf{v}_{y}-\mathbf{v}_{r,y}\|_2^2  \right\}$ & 1.0 & 1.0 \\
    
    Ang. Vel. tracking & $\exp \left\{-4 \left\|\omega_{\text {yaw }}-\omega_{\text{r,yaw }}\right\|^2\right\}$ & 2.0 & 1.0 \\
    Base height tracking & $\exp \left\{-4 \left\|h_{\text {t}}-h_{\text {r,t}}\right\|^2\right\}$ &2.0&2.0\\
    Lin. Vel. z & $v_{r,z}^2$ & -0.5 & -0.5 \\
    Ang. Vel. xy & $\|\boldsymbol{\omega}_{r,x y}\|_2^2$ & -0.025 & -0.025 \\
    Orientation & $\|\mathbf{g}_{x}\|_2^2 + \|\mathbf{g}_{y}\|_2^2$ & -1.5 & -1.5 \\
    Action rate & $\|\mathbf{a}_t-\mathbf{a}_{t-1}\|_2^2$ & -0.01 & -0.01 \\
    Hip joint deviation & $\sum\limits_{\text{hip joints}}|\theta_{i} - \theta^{default}_{i}|^{2}$ & -0.2 & -0.5 \\
    Ankle joint deviation & $\sum\limits_{\text{ankle joints}}|\theta_{i} - \theta^{default}_{i}|^{2}$ & -0.5 & -0.75 \\
    Squat knee & $-\|(h_{r,t}-h_{t})\times(\frac{q_{knee,t}-q_{knee,min}}{q_{knee,max}-q_{knee,min}}-\frac{1}{2})\|$ & -0.75 & -0.75 \\
    Dof Acc. & $\sum\limits_{\text{all joints}} \frac{\|\dot q_{t,i}-\dot q_{t-1,i}\|^2}{dt}$ & $-2.5 \times 10^{-7}$ &  $-2.5 \times 10^{-7}$\\
    Dof pos limits & $\sum\limits_{\text{all joints}} \text{out}_{i}$ & -2.0 & -2.0 \\
    Feet air time & $\mathbf{1}_{\{\text{first contact}\}}(T_{air}-0.5)$ & 0.05 & 0.05\\
    Feet clearance & $\sum\limits_{feet} \left(p_{z}^{\text {target}}-p_{z}^{i}\right)^2 \cdot v_{xy}^{i}$ & -0.25 & -0.25 \\
    Feet lateral distance & $|y_{\text{left foot}}^{B} - y_{\text{right foot}}^{B}| - d_{min}$ & 0.5 & 0.5 \\
    Knee lateral distance & $|y_{\text{left knee}}^{B} - y_{\text{right knee}}^{B}| - d_{min}$ & 1.0 & 1.0 \\
    Feet ground parallel & $\sum\limits_{feet}Var(H_i)$ & -2.0 & -2.0 \\
    Feet parallel & $Var(D)$ & -3.0 & -3.0 \\
    Smoothness & $\|\mathbf{a}_t-2 \mathbf{a}_{t-1}+\mathbf{a}_{t-2}\|_2^2$ & -0.05 & -0.05 \\
    Joint power & $\frac{|\boldsymbol{\tau} \| \dot{\boldsymbol{\theta}}|^{T}}{\|\mathbf{v}\|_2^2 + 0.2 * \|\boldsymbol{\omega}\|_2^2}$ & $-2.0 \times 10^{-5}$ & $-2.0 \times 10^{-5}$ \\
    Feet stumble & $\mathbf{1}\left\{\exists i,\left|\mathbf{F}_i^{x y}\right|>3\left|F_i^z\right|\right\}$ & -1.5 & -1.5 \\
    Torques & $\sum\limits_{\text{all joints}} |\frac{\tau_i}{kp_{i}}|_{2}^{2}$ & $-2.5 \times 10^{-6}$ & $-2.5 \times 10^{-6}$ \\
    Dof Vel. & $\sum\limits_{\text{all joints}} \dot{\theta}_{i}|_{2}^{2}$ & $-1 \times 10^{-4}$ & $-1 \times 10^{-4}$ \\
    Dof Vel. limit & $\sum\limits_{\text{all joints}} RELU(\hat{\theta}_{i} - \hat{\theta}^{max}_{i})$ & $-2\times 10^{-3}$ & $-2\times 10^{-3}$ \\
    Torque limits & $\sum\limits_{\text{all joints}} RELU(\hat{\tau}_{i} - \hat{\tau}^{max}_{i})$ & -0.1 & -0.2 \\
    No fly & $\mathbf{1}\{\text{only one feet on ground}\}$ & 0.75 & 0.5 \\
    Joint tracking error & $\sum\limits_{\text{all joints}}|\theta_{i} - \theta^{target}_{i}|^{2}$ & -0.1 & -0.25 \\
    Feet slip & $\sum\limits_{feet}\left|\mathbf{v}_i^{\text {toot }}\right| * \sim \mathbf{1}_{\text {new contact }}$ & -0.25 & -0.25 \\
    Feet contact force & $\sum\limits_{feet} RELU(F_{i}^{z} - F_{th})$ & $-2.5 \times 10^{-4}$ & $-2.5 \times 10^{-4}$ \\
    Contact momentum & $\sum\limits_{feet}|v_{i}^{z} * F_{i}^{z}|$ & $2.5 \times 10^{-4}$ & $2.5 \times 10^{-4}$ \\
    Action vanish & $\sum\limits_{\text{all joints}}(\max\{0, a_{i,t}-a_{i,max}\}+\min\{0, a_{i,min}-a_{i,t}\})$&-1.0 & -1.0\\
    Stand still & $Num_{\{feet\,not\,on\,ground\}}\times \mathbf{1}_{stand\,still}$ &-0.15 & -0.2\\
    \bottomrule[1.5pt]
    \end{tabular}
    \label{tab:reward}
\end{table}
\subsection{Domain Randomization}
\label{appendix:Random}
To improve the robustness of the trained policy, we employ domain randomization to simulate several kinds of random noises that may occur while deploying in the real world. The terms used for randomization, along with their descriptions and ranges, are listed in \cref{tab:domain}. Specifically, we introduce a term to randomize the mass of the hands, enhancing the robot's capability to hold objects effectively. The ranges for the Unitree G1 and Fourier GR-1 are the same.
\begin{table}[!ht]
    \centering
    \caption{Randomization Terms, Description, and Ranges}
    \begin{tabular}{llc}
    \toprule[1.5pt] Term & Description & Ranges \\
    \midrule[1.5pt] 
    Actuation offset $(N\cdot m)$ & Random torque offsets applied to the computed motor torques & $[-0.05, 0.05]$ \\
    Torso payload mass $(Kg)$ & Additional random mass attached to the torso and hand links & $[-5.00, 10.00]$ \\
    Hand payload mass $(Kg)$ & Additional random mass attached to the hand links & $[-0.10, 0.30]$ \\
    CoM displacement $(m)$ & Random offsets applied to the center of mass (CoM) position of the torso link & $[-0.1, 0.1]$ \\
    Link mass $(-)$ & Random scaling factors applied to the masses of the robot's links & $[0.80, 1.20]$ \\
    Friction coefficient $(-)$ & Random friction coefficients applied to the robot's links & $[0.10, 2.00]$ \\
    Restitution $(-)$ & Random restitution coefficients applied to the robot's links & $[0.00, 1.00]$ \\
    $K_p$ $(N/rad)$ & Random scaling factors applied to the proportional gain ($K_p$) of the robot's joints & $[0.90, 1.10]$ \\
    $K_d$ $(N/(m/s))$ & Random scaling factors applied to the derivative gain ($K_d$) of the robot's joints & $[0.90, 1.10]$ \\
    Initial joint pos scale $(-)$ & Random scaling factors applied to the initial positions of the robot's joints & $[0.80, 1.20]$ \\
    Initial joint pos offset $(rad)$ & Random offsets added to the initial positions of the robot's joints & $[-0.10,0.10]$ \\
    Push robot $(m/s)$ & Random x and y velocities applied to the robot to simulate external pushes & $[-0.50, 0.50]$ \\
    Dof pos obs $(rad)$& Random dof velocity added to the observation of joint positions & $[-0.02, 0.02]$ \\
    Dof vel obs $(rad/s)$& Random dof velocity added to the observation of joint velocities & $[-2.00, 2.00]$  \\
    Ang vel obs $(rad/s)$& Random dof velocity added to the observation of body angular velocities & $[-0.50, 0.50]$ \\
    Gravity obs $(m/s^2)$& Random dof velocity added to the observation of gravities projected to robot's body frame & $[-0.05, 0.05]$ \\
    \bottomrule[1.5pt]
    \end{tabular}
    \label{tab:domain}
\end{table}
\subsection{Other Key Parameters}
\label{appendix:OtherKey}
We list other key parameters used to train the Unitree G1 and Fourier GR-1 in \cref{tab:para}. The same settings are applied for both training and evaluation. Additionally, we adjust the base height target value when the environment is used to train squatting; otherwise, the robot is required to track a constant height value while walking. Terms marked with $*$ indicate that exceeding the specified range will result in penalties through corresponding rewards.
\begin{table}[!ht]
    \centering
    \caption{Key Parameters Used to Train Robots}
    \begin{tabular}{lcc}
    \toprule[1.5pt] Term & Unitree G1 & Fourier GR-1 \\
    \midrule[1.5pt] 
    Height target while walking $(m)$& 0.74 & 0.90 \\
    X Lin. Vel. range $(m/x)$ & $[-0.80, 1.20]$ & $[-0.80, 1.20]$ \\
    Y Lin. Vel. range $(m/s)$& $[-0.50, 0.50]$ & $[-0.80, 0.80]$ \\
    Yaw Ang. Vel. range $(rad/s)$& $[-0.80, 0.80]$ & $[-1.00, 1.00]$ \\
    Squat height range $(m)$& $[-0.24, 0.74]$ & $[-0.30, 0.90]$ \\
    Soft dof pos limit scale * $(-)$& 0.975 & 0.975 \\
    Soft dof vel limit scale * $(-)$& 0.80 & 0.80 \\
    Soft dof torque limit scale * $(-)$& 0.95 & 0.95 \\
    Max contact force * $(N)$& 400.00 & 500.00 \\
    
    Least feet distance * $(m)$& 0.20 & 0.20 \\
    Least knee distance * $(m)$& 0.20 & 0.20 \\
    Most feet distance * $(m)$& 0.35 & 0.40 \\
    Most knee distance * $(m)$& 0.35 & 0.40 \\
    Clearance height target * $(m)$& 0.14 & 0.15 \\
    Push interval $(s)$& 4.00& 4.00\\
    Upper-body poses resampling interval $(s)$& 1.00& 1.00\\
    Commands resampling interval $(s)$& 4.00& 4.00\\
    \bottomrule[1.5pt]
    \end{tabular}
    \label{tab:para}
\end{table}
\subsection{Function Visualization}
\label{appendix:FunVisual}
\begin{figure}[!ht]
  \centering
  \includegraphics[width=0.8\textwidth]{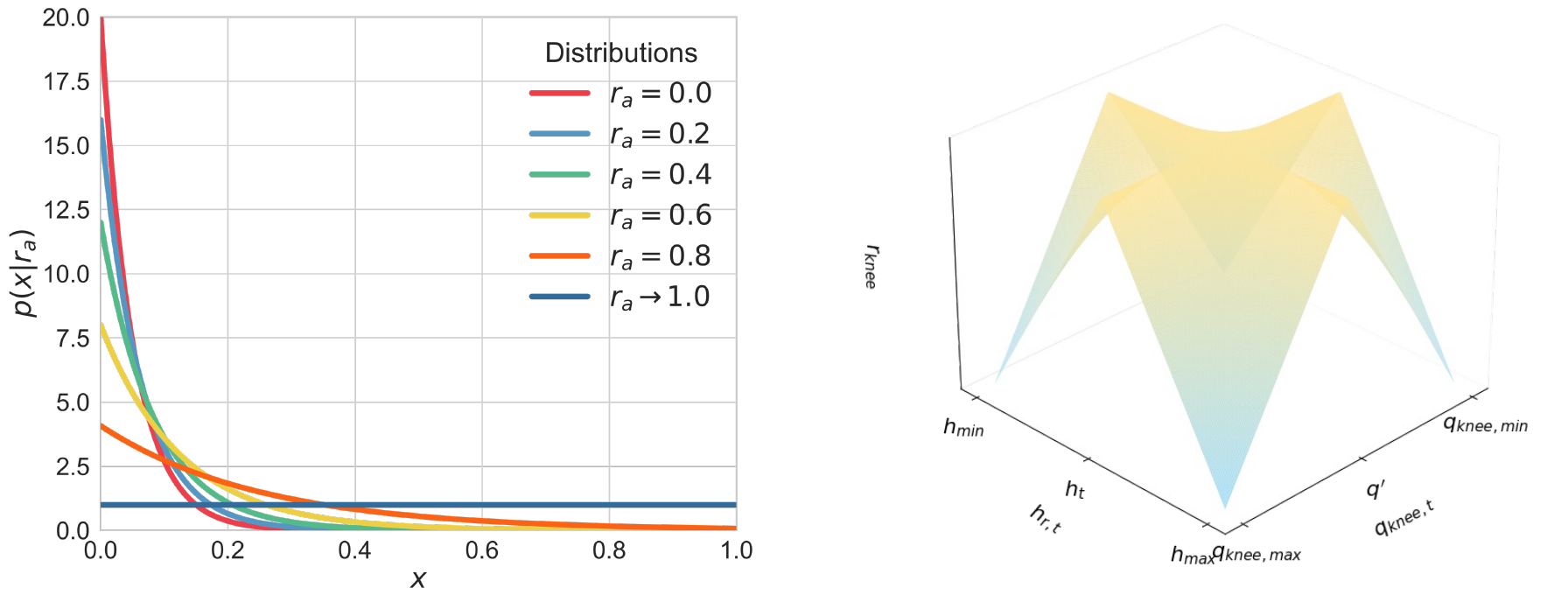}
  \caption{Visualization of proposed functions. \textcolor{mycolor}{Left:} Visualization of $p(x|r_a)$ in \cref{equa:sample}. \textcolor{mycolor}{Right:} Visualization of $r_{knee}$ in \cref{eq:rknee}.}
  \label{fig:visual}
\end{figure}

For better understanding of the equations we proposed in \cref{equa:sample} and \cref{eq:rknee}, we visualize them in \cref{fig:visual}. As shown in the left figure, $p(x|r_a)$ can take any value in the range $[0,1]$ when $r_a \in [0,1)$. When $r_a$ is small, it is more likely to take smaller values of $x$. As $r_a$ increases, the probability of taking larger values of $x$ also increases. When $r_a \rightarrow 1$, the entire distribution becomes $\mathcal{U}(0,1)$. In the right figure, we can observe that regardless of the position of $(h, q)$, $r_{knee}$ encourages $q_{knee,t}$ to change in a direction that brings $h_{r,t}$ closer to $h_{t}$. This achieves the goal of guiding the robot to track the base height by either bending or straightening its knees.
\subsection{Terrain Traverse}
\label{appendix:Terrain}
\begin{figure}[!ht]
  \centering
  \includegraphics[width=0.98\textwidth]{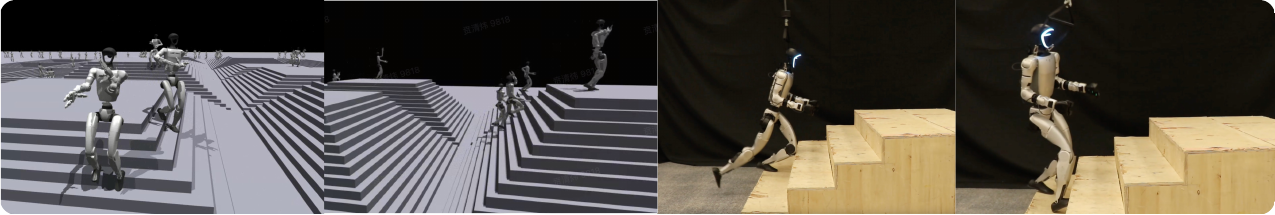}
  \caption{Training robots to traverse stairs. \textcolor{mycolor}{Left:} Training in simulation; \textcolor{mycolor}{Right:} Deployment in the real world.}
  \label{fig:terrain}
\end{figure}

In order to expand the feasible workspace of robots, we integrate our training framework with a previous humanoid locomotion method called PIM \cite{long2024learninghumanoidlocomotionperceptive} to enable our robots to traverse stairs. As shown in the left figure of \cref{fig:terrain}, we successfully train the Unitree G1 in Isaac Gym to traverse high stairs. However, when deploying the trained policy in the real world, as shown in the right figure of \cref{fig:terrain}, the robot struggles to walk stably and collides with the stair, despite eventually stepping onto it. This instability arises because the head of the Unitree G1 cannot remain fixed, causing movement of the LiDAR mounted on it. Additionally, the elevation map acquisition method used by PIM lacks high resolution, further exacerbating the sim2real gap. In the future, we will explore methods to truly enable our robots to traverse any terrain effectively.

\section{Hardware System Details}
\label{appendix:Hardware}
In \cref{fig:overview}, we present the hardware system design framework of HOMIE, which comprises three integral components: isomorphic exoskeleton arms, a pair of motion-sensing gloves, and a pedal. The primary structural elements of these components are fabricated using 3D printing technology with PLA basic material. \red{PLA provides adequate strength at a low cost, and 3D-printed components can be easily built and modified. Building hardware with other materials is feasible using our CAD models.}
\subsection{Isomorphic Exoskeleton Details}
\label{appendix:Exoskeleton}
\begin{figure}[!ht]
  \centering
  \includegraphics[width=1.0\textwidth]{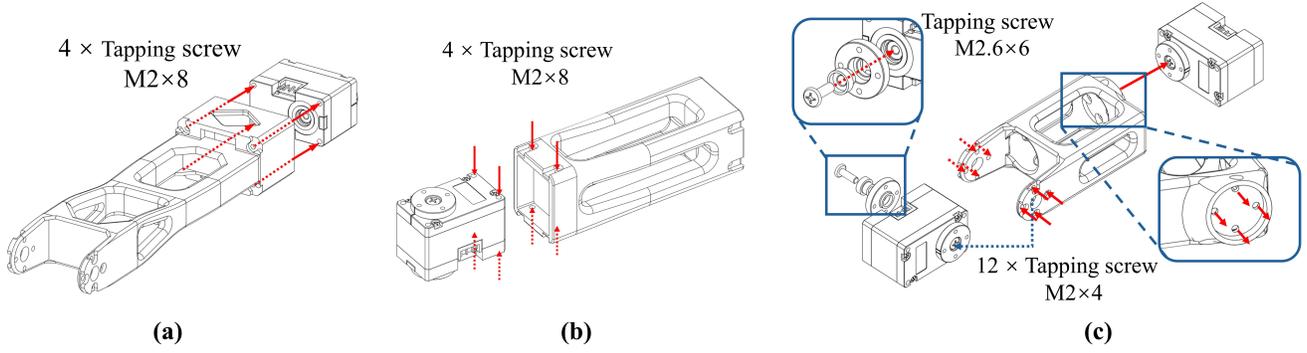}
  \caption{The assembly methods of the servos and structural components, along with the screw requirements, are as follows: \textcolor{mycolor}{(a)}, \textcolor{mycolor}{(b:)} The structural components are directly assembled and fixed to the servo body; \textcolor{mycolor}{(c:)} The structural components are assembled and fixed to the servos via one or two servo disks, respectively.}
  \label{fig:assemble}
\end{figure}

\begin{figure}[!ht]
  \centering
  \includegraphics[width=0.94\textwidth]{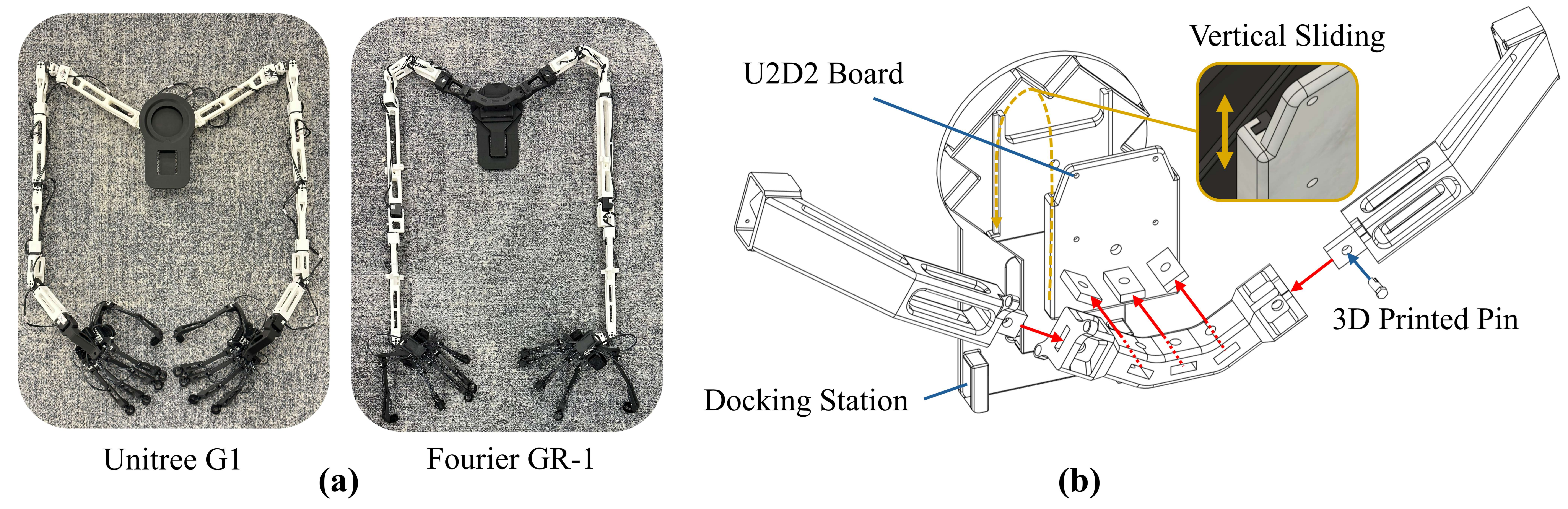}
  \caption{\textcolor{mycolor}{(a:)} Physical models of two different Humanoids' Isomorphic Exoskeletons, equipped with servos, back connectors, and motion-sensing gloves; \textcolor{mycolor}{(b:)} Schematic diagram of the back connector assembly and functionality, where the connectors can be fixed using a dovetail structure and plugs. The U2D2 board and docking station are external physical components.}
  \label{fig:back}
\end{figure}

The Isomorphic Exoskeleton adopts a hollowed-out and mortise-and-tenon design, which not only ensure structural integrity but also significantly reduce the overall weight and assembly complexity, as well as facilitate the routing of servo motor connections. The structural components are fixed to the servos through four types of connection methods: two methods that directly connect to the servo body \cref{fig:assemble}\textcolor{mycolor}{(a)}, \textcolor{mycolor}{(b)} and two methods that connect to the servo disks \cref{fig:assemble}\textcolor{mycolor}{(c)} To further enhance stability and strength, additional servo disks are installed on the opposite side of the servos at certain joints, allowing the structural components to connect directly to the servo disks on both sides, as illustrated in \cref{fig:assemble}\textcolor{mycolor}{(c)}. We present the physical models of the Isomorphic Exoskeletons adapted for the  Unitree G1 and Fourier GR1 in \cref{fig:back}\textcolor{mycolor}{(a)}. Due to the different configurations of the Humanoid, there are significant structural differences in the wrist and shoulder components between the two sets of Isomorphic Exoskeletons, while the other components and their usage remain identical. The two sets of Exoskeletons share the same back connector, which integrates functionalities for operator attachment, docking station fixation, U2D2 placement, and bilateral arm linkage, as illustrated in \cref{fig:back}\textcolor{mycolor}{(b)}.

\subsection{Motion-sensing Glovesc Details}
\label{sec:Glovesc}
\begin{figure}[!ht]
  \centering
  \includegraphics[width=0.6\textwidth]{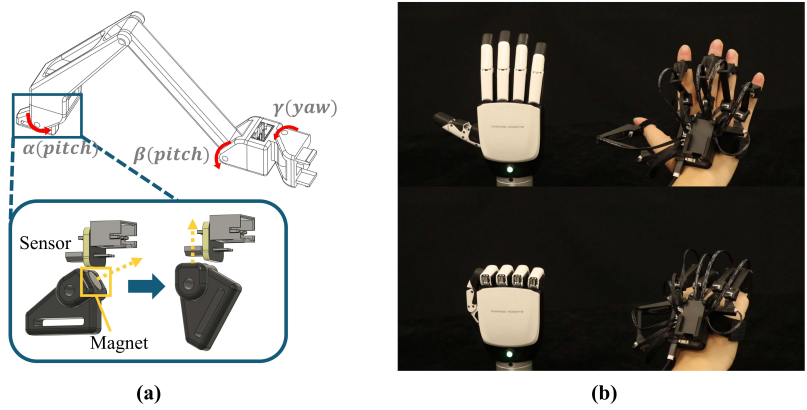}
  \caption{\textcolor{mycolor}{(a):} The schematic diagrams of the three degrees of freedom for each finger and the rotation of the magnet affecting the magnetic field direction. \textcolor{mycolor}{(b):} Physical images of the Inspire Hands actual device in both open and clenched fist states.}
  \label{fig:glove}
\end{figure}

The Unitree G1 is equipped with the Unitree Dex3-1, a three-fingered dexterous hand with 7 DoF(three in the thumb and two each in the middle and index fingers). Our motion-sensing gloves can track up to 15 DoF, enabling direct mapping of thumb, middle finger, and index finger movements to the corresponding fingers on the Unitree Dex3-1. The motion-sensing gloves are secured to the palm via a length-adjustable elastic strap and connected to the fingertips through five smaller length-adjustable elastic straps, which facilitate finger fixation and critical angle mapping, ensuring adaptability to operators with varying hand sizes. A microcontroller is embedded within the palm section of the gloves, featuring exposed ports that allow direct connection to the 15 Hall sensor modules located at the fingers, as depicted in \cref{fig:glove}\textcolor{mycolor}{(a)}.
For the joint mapping angle range and acquisition accuracy of each finger, we have listed the data in \cref{tab:glove_range}. Each finger of the glove has three degrees of freedom, which are shown in \cref{fig:glove}\textcolor{mycolor}{(a)}, namely the pitch motion of the fingertip ($\alpha$), the pitch motion of the finger pad ($\beta$), and the yaw motion of the finger pad ($\gamma$). Due to differences in the structural length of the thumb, the pinky, and the other three fingers, we have divided them into three parts. It should be noted that the angular movement of the finger joints does not exhibit a significant linear relationship with the changes in the Hall sensor readings caused by the induced magnetic field variation. This is related to the positioning of the magnets and Hall sensors, as well as the structural design of the gloves. In our motion-sensing gloves design, the relationship between the two follows an exponential pattern within their transformation range, especially in the pitch motion of the fingers and finger pads, from open to fist. Furthermore, our gloves have undergone control testing on the Inspire Dexterous Hands RH56DFTP actual device, illustrated in \cref{fig:glove}\textcolor{mycolor}{(b)}.

\begin{table}[!ht]
    \centering
    \caption{15 DoF Joint Mapping Angle Range and Acquisition Accuracy. (Acc:accuracy)}
    \begin{tabular}{llccc}
    \toprule[1.5pt] Term & Description & Angle Range & Acquisition Range & Acquisition Acc.\\
    \midrule[1.5pt] 
     $\alpha_{thumb}$ & Pitch motion of the thumb tip & 65\text{\textdegree} & 528 units & 0.123\( \text{\textdegree} / \text{unit} \) \\
     $\beta_{thumb}$ & Pitch motion of the thumb pad & 100\text{\textdegree} & 1024 units & 0.098\( \text{\textdegree} / \text{unit} \) \\
     $\gamma_{thumb}$ & Yaw motion of the thumb pad & 90\text{\textdegree} & 832 units & 0.108\( \text{\textdegree} / \text{unit} \) \\
     $\alpha_{pinky}$ & Pitch motion of the pinky tip & 70\text{\textdegree} & 880 units & 0.080\( \text{\textdegree} / \text{unit} \) \\
     $\beta_{pinky}$ & Pitch motion of the pinky pad & 90\text{\textdegree} & 1136 units & 0.079\( \text{\textdegree} / \text{unit} \) \\
     $\gamma_{pinky}$ & Yaw motion of the pinky pad & 45\text{\textdegree} & 416 units & 0.108\( \text{\textdegree} / \text{unit} \) \\
     $\alpha_{other}$ & Pitch motion of the index, middle, and ring finger tips & 70\text{\textdegree} & 928 units & 0.075\( \text{\textdegree} / \text{unit} \) \\
     $\beta_{other}$ & Pitch motion of the index, middle, and ring finger pads & 90\text{\textdegree} & 1072 units & 0.088\( \text{\textdegree} / \text{unit} \) \\
     $\gamma_{other}$ & Yaw motion of the index, middle, and ring finger pads & 40\text{\textdegree} & 512 units & 0.078\( \text{\textdegree} / \text{unit} \) \\
     
    \bottomrule[1.5pt]
    \end{tabular}
    \label{tab:glove_range}
\end{table}

\subsection{Foot Pedal Details}
\label{appendix:Pedal}
\begin{figure}[!ht]
  \centering
  \includegraphics[width=0.8\textwidth]{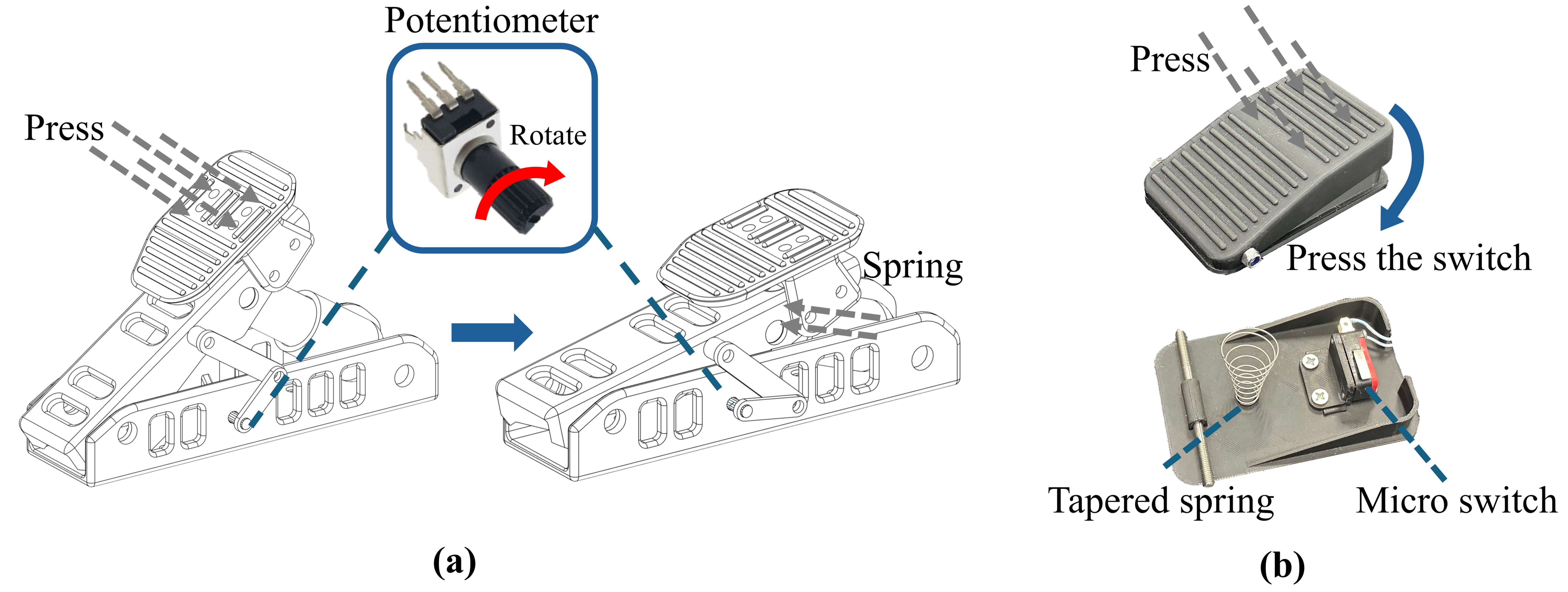}
  \caption{\textcolor{mycolor}{(a):} Schematic diagram of the small pedal principle, where the operator's foot press drives the potentiometer to rotate. \textcolor{mycolor}{(b):} Schematic diagram of the mode-switching buttons, where the operator's foot press changes the state of the micro switch.}
  \label{fig:pedal_app}
\end{figure}

The foot pedal consists of three small pedals and two mode-switching buttons, all fixed onto a large base plate, as shown in \cref{fig:pedal}. The operator can press the small pedals, which cause the structural components to rotate, thereby driving the potentiometer at the bottom to rotate. The spring within the structure ensures that when the operator releases the pedal, it springs back, returning to the initial position, as shown in \cref{fig:pedal_app}\textcolor{mycolor}{(a)}. The potentiometer we use, model 0932, has a range of angular movement of 270\text{\textdegree}, with the actual pedal movement range being 40\text{\textdegree}. As for the mode-switching buttons, the operator can press the buttons on the surface, which will cause a change in the high and low levels of the micro switch, as shown in \cref{fig:pedal_app}\textcolor{mycolor}{(b)}. The function of the tapered spring is the same as the spring in the small pedals.

\section{Deployment Details}
\label{appendix:deploydetail}

\subsection{System Deployment}

\begin{figure}[!ht]
  \centering
  \includegraphics[width=0.8\textwidth]{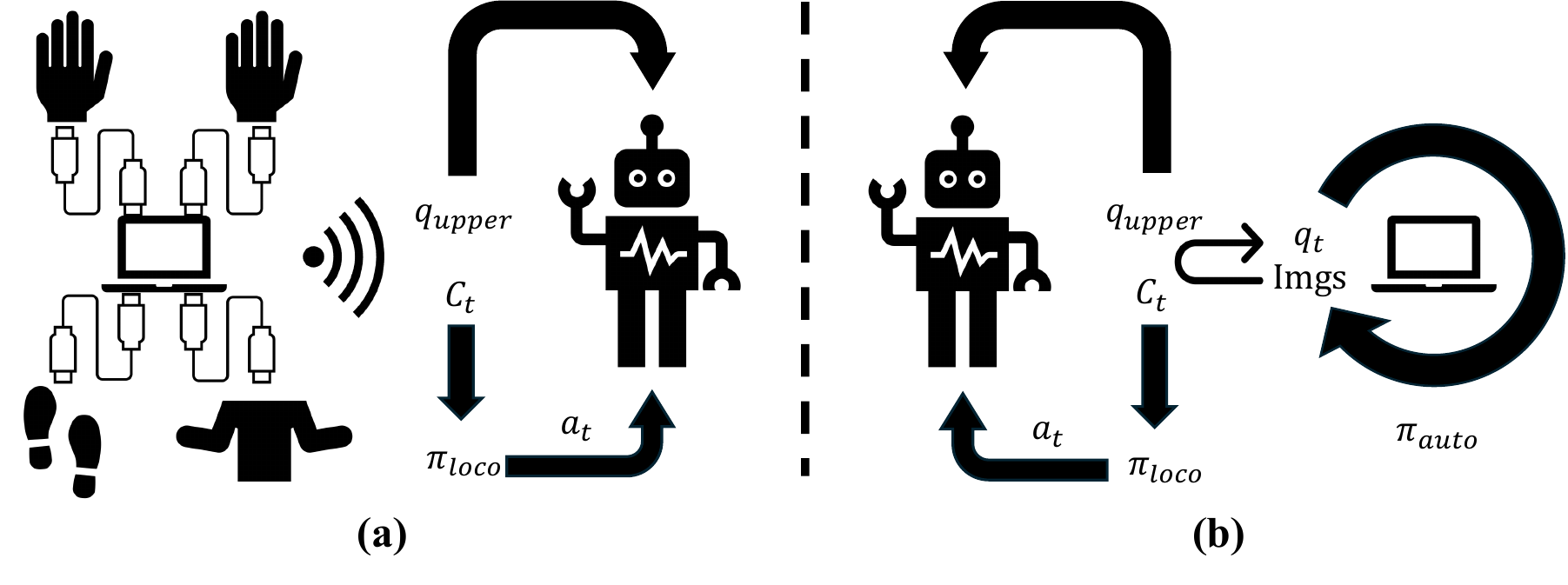}
  \caption{Deployment of our system with cockpit and autonomous policy. \textcolor{mycolor}{(a):} Deployment with robot controlled by isomorphic exoskeleton cockpit. \textcolor{mycolor}{(b):} Deployment with robot controlled by autonomous policy $\pi_{auto}$.}
  \label{fig:deploy}
\end{figure}

We deploy our trained policy $\pi_{loco}$ directly onto the Unitree G1’s onboard computing unit—a Nvidia Jetson Orin capable of 275 TOPS—allowing $\pi_{loco}$ to run at 50\,Hz using the robot’s state information to control walking and squatting, matching the frequency used during Isaac Gym training. We use an isomorphic exoskeleton-based approach to control the robot, as shown in \cref{fig:deploy}(b). A CPU-only host computer connects via four data lines to the isomorphic arm, left and right gloves, and the pedal’s microcontroller, reading real-time data. It then transmits $q_{upper}$ and $C_t$ to the G1 over Wi-Fi via TCP, enabling the robot to set upper-body poses and compute lower-body actions $a_t$ through $\pi_{loco}$ for full-body control, while simultaneously returning real-time images to the host. The D455 camera provides $640\times480$ images at roughly 30Hz over TCP. Due to hardware constraints and TCP network limitations, the G1 cannot directly process high-frequency data. In our deployment, we therefore update the arm joint position targets at 10Hz and interpolate these targets 50 times to drive the robot’s upper body smoothly. However, if a robot can accept higher-frequency control signals, our system can support an update frequency over 200Hz.

\subsection{IL Deployment}
Once we train the policy $\pi_{auto}$ using the method described in \cref{sec:iltraining}, we deploy it as illustrated in \cref{fig:deploy}(b). We connect the Unitree G1 to a host equipped with an Nvidia RTX 4080 GPU via an Ethernet cable, enabling wired TCP communication. This wired setup provides the faster data transfer needed for transmitting images. The runtime involves two processes on the G1, labeled \textcolor{mycolor}{1} and \textcolor{mycolor}{2}, and one process on the host, labeled \textcolor{mycolor}{3}. Process \textcolor{mycolor}{1} captures images from the D455 camera mounted on the G1’s head and from the D435 cameras on its arms, along with the robot’s joint information $q_{t,i}$. This data is then sent to the host. Process \textcolor{mycolor}{3} receives the inputs, performs inference of $\pi_{auto}$ to compute $C_t$ and $q_{upper}$, and returns these results to the G1. Finally, process \textcolor{mycolor}{2} controls the robot’s motion using the inferred commands. The entire loop runs at a frequency of 10Hz.

\subsection{Simulation Deployment}
We train the policies in Isaac Gym, which is sufficient for training locomotion policies but lacks the capability to simulate realistic scenes. Therefore, we employ a sim2sim process to transfer our policies to Isaac Sim. The core of this process involves aligning the joint order and quaternion conventions between the two platforms. In Isaac Gym, the joint order follows depth-first ordering, and quaternions are formatted as xyzw. In contrast, Isaac Sim uses breadth-first ordering for joints and wxyz for quaternions~\cite{mittal2023orbit}. As a result, both the neural network's observation $O_{t-5:t}$ and the computed action $a_t$ must undergo corresponding order adjustments. To simulate the real-world camera perspective in the simulation, we directly add a Camera to the robot's USD file and place its prim path under the prim path of the link to which it is bound. This ensures that the camera moves along with the corresponding link during motion in the simulation. The camera parameters, such as resolution and focal length, are configured to match those used in the real world.

\section{\red{User Study}}
\subsection{\red{User Study Raw Data}}
\label{sec:raw_data}
\red{We list raw data got in user study in~\cref{tab:user}. Our testers represented a diverse spectrum of heights, body types, and genders performing handover tasks. All participants were equipped with identical exoskeleton hardware throughout trials. Despite anthropometric variations from the system's design specifications, every tester achieved expert-level proficiency within minimal practice time, demonstrating the system's strong adaptability to diverse body types and inherent user-friendliness.}
\begin{table}[!ht]
    \centering
    \caption{User Study Raw Data}
    \begin{tabular}{cccccccccc}
    \toprule[1.5pt] User ID & User Height (cm) & User Weight (kg) & User Gener & Time1(s) & Time2(s) & Time3(s) & Time4(s) & Time5(s) \\
    \midrule[1.5pt] 
     1 & 185 & 71 & Male & 34 & 28 & 15 & 16 & 9 \\
     2 & 185 & 90 & Male & 19 & 16 & 13 & 7 & 6\\
     3 & 160 & 55 & Female & 19 & 28 & 14 & 17 & 12 \\
     4 & 170 & 63 & Male & 29 & 21 & 14 & 13 & 6 \\
     5 & 175 & 80 & Male & 23 & 22 & 15 & 12 & 7 \\
     Average & - & - & - & 24.8 & 23.0 & 14.2 & 13.0 & 8.0 \\
    \bottomrule[1.5pt]
    \end{tabular}
    \label{tab:user}
\end{table}

\subsection{\red{Teaching Procedure}}
\label{sec:teaching}
\red{The standardized onboarding protocol comprises six key phases as shown below, with operational objectives paired with corresponding verbal guidance examples:}
\begin{itemize}
    \item\red{Customized exoskeleton fitting with adjustable straps for user-centric anthropometric adaptation: ``Please wear the exoskeleton backpack-style by fastening the two back straps, then secure the finger straps on each fingertip. Adjust strap tightness according to your comfort level.''}
    
    \item\red{Ergonomic positioning featuring seat-height optimization and pedal placement calibration: ``Please sit on the chair and adjust the pedal distance to your legs for optimal comfort. Try pressing the pedal to experience the operation.''}
    
    \item\red{Systematic briefing on pedal control layout with functional mapping visualization: ``The white center pedal controls robot squatting. The large right pedal controls forward/backward movement - press the small right pedal to toggle direction. The large left pedal controls turning, with its small pedal switching left/right orientation.''}
    
    \item\red{Motion-sensing glove familiarization through bidirectional mapping demonstrations: ``The dexterous hand control program is now active. Move your fingers to experience how each robotic finger's 2-3 joints correspond to your glove's knuckle joints and fingertip sensors.''}
    
    \item\red{Immersive exoskeleton-arm coordination training via real-time manipulation exercises: ``The full-body control program is activated. Move your arms to observe mirrored robot movements. Use pedals for squatting or locomotion commands.''}
    
    \item\red{Structured task execution guidance for teleoperation mastery: ``You may now combine exoskeleton movements with pedal operations to complete loco-manipulation tasks through the humanoid robot.''}
\end{itemize}

\red{The entire protocol is typically completed within \textbf{5 minutes} from initial fitting to task execution.}
\end{appendices}

\end{document}